\documentclass[conference]{IEEEtran}
\IEEEoverridecommandlockouts

\usepackage[numbers,sort&compress]{natbib}
\usepackage{tikz}
\usepackage{amsmath}
\usepackage{amssymb}
\usepackage{pifont}
\usepackage{bbm}
\usepackage{subfig}
\usepackage{booktabs}
\usepackage{url}
\usepackage{hyperref}
\usepackage{paralist}
\usepackage{float}
\usepackage{tabularx}
\usepackage{xparse}
\usepackage[indLines=false]{algpseudocodex}
\usepackage{algorithm}
\usepackage{xcolor}
\usepackage{colortbl}  
\usepackage{xspace}
\usepackage{balance}
\usepackage{nicefrac}
\usepackage{enumitem}
\usepackage[nameinlink,capitalise]{cleveref}
\PassOptionsToPackage{table}{xcolor}

\NewDocumentCommand{\statcirc}{ O{#2} m }{%
    \begin{tikzpicture}
    \node[circle,minimum width=2mm,draw,fill=#1] {};
    \fill[#2] (0,0) circle (1.0ex); 
    \fill[#1] (0,0) -- (90:1ex) arc (90:270:1ex) -- cycle; 
    \end{tikzpicture}
}

\algblockdefx[game]{game}{EndGame}[2]{\textbf{experiment} \textsc{#1}(#2)}{\algpx@endIndent}
\algtext*{EndGame}

\let\oldReturn\Return
\renewcommand{\Return}{\State\oldReturn}

\definecolor{redbg}{RGB}{254,241,240}
\definecolor{redoutline}{RGB}{252,163,152}
\definecolor{redtext}{RGB}{207,24,34}
\DeclareRobustCommand*{\rectangled}[1]{%
  \tikz[baseline=(char.base)]\node[anchor=south west, draw, rectangle, thick, rounded corners=0.2mm, inner sep=2pt, fill=redbg, draw=redoutline,text=redtext](char){#1} ;}

\DeclareMathOperator*{\argmax}{arg\,max}
\DeclareMathOperator*{\argmin}{arg\,min}
\DeclareMathOperator{\EX}{\mathbb{E}}

\newtheorem{definition}{Definition}

\newcommand{\iid}{i.i.d.\@\xspace}
\newcommand{\eg}{e.g.\@\xspace}
\newcommand{\ie}{i.e.\@\xspace}
\newcommand{\wrt}{w.r.t.\@\xspace}

\begin{document}

\title{Analyzing Leakage of Personally Identifiable Information in Language Models}

 \author{%
   \IEEEauthorblockN{%
     Nils Lukas\IEEEauthorrefmark{1}\textsuperscript{\textsection},
     Ahmed Salem\IEEEauthorrefmark{2},
     Robert Sim\IEEEauthorrefmark{2},
     Shruti Tople\IEEEauthorrefmark{2},
     Lukas Wutschitz\IEEEauthorrefmark{2} and
     Santiago Zanella-B{\'e}guelin\IEEEauthorrefmark{2}
   }%
   \IEEEauthorblockN{\IEEEauthorrefmark{1}University of Waterloo, \IEEEauthorrefmark{2}Microsoft}%
   \IEEEauthorblockN{
     nlukas@uwaterloo.ca, \{t-salemahmed, rsim, shruti.tople, lukas.wutschitz, santiago\}@microsoft.com}
 }
    
\maketitle
 \begingroup\renewcommand\thefootnote{\textsection}
 \footnotetext{Part of this work was done during an internship at Microsoft Research.}
 \endgroup
 \begingroup\renewcommand\thefootnote{\textsuperscript{\ddag}}
 \footnotetext{\footnotesize{To cite this work, please refer to the full publication~\cite{lukas2023analyzing} in IEEE Security and Privacy (S\&P) 2023.}}.
 \endgroup

\begin{abstract}
Language Models (LMs) have been shown to leak information about training data through sentence-level membership inference and reconstruction attacks.
Understanding the risk of LMs leaking Personally Identifiable Information (PII) has received less attention, which can be attributed to the false assumption that dataset curation techniques such as scrubbing are sufficient to prevent PII leakage. 
Scrubbing techniques reduce but do not prevent the risk of PII leakage: in practice scrubbing is imperfect and must balance the trade-off between minimizing disclosure and preserving the utility of the dataset. 
On the other hand, it is unclear to which extent algorithmic defenses such as differential privacy, designed to guarantee sentence- or user-level privacy, prevent PII disclosure.
In this work, we introduce rigorous game-based definitions for three types of PII leakage via black-box extraction, inference, and reconstruction attacks with only API access to an LM.
We empirically evaluate the attacks against GPT-2 models fine-tuned with and without defenses in three domains: case law, health care, and e-mails.  
Our main contributions are
\begin{inparaenum}[(i)]
\item novel attacks that can extract up to 10$\times$ more PII sequences than existing attacks,
\item showing that sentence-level differential privacy reduces the risk of PII disclosure but still leaks about 3\% of PII sequences, and
\item a subtle connection between record-level membership inference and PII reconstruction.
Code to reproduce all experiments in the paper is available at \url{https://github.com/microsoft/analysing_pii_leakage}.
\end{inparaenum}
\end{abstract}

\section{Introduction}
\label{sec:intro}
Language Models (LMs) are fundamental to many natural language processing tasks~\cite{hoang2019efficient,nakano2021webgpt}. 
State-of-the-art LMs scale to trillions of parameters~\cite{fedus2021switch} and are pre-trained on large text corpora (\eg, 700GB~\cite{raffel2020exploring}).
Pre-trained LMs are adapted to downstream tasks by fine-tuning on domain-specific datasets such as human dialogs~\cite{budzianowski2019hello} or clinical health data~\cite{vakili2022downstream} which may contain private information. 

Memorization is a privacy concern in LMs~\cite{carlini2021extracting}. 
The threat is that an attacker learns \emph{by whom} the training data was provided, known as membership inference~\cite{shokri2017membership,jagannatha2021membership, mireshghallah2022quantifying, mireshghallah2022memorization} and \emph{about whom} it contains information, known as data extraction~\cite{carlini2021extracting, stock2022defending,zhang2022text,carlini2022quantifying,ippolito2022preventing}. 
These two categories can be disjoint but associations in the latter can be used to infer information about the former. 
For LMs, data extraction is a significant threat in practice since attackers with black-box API access can extract at least 1\% of the training data~\cite{carlini2022quantifying}. 

Existing work focuses on finding a lower bound on \emph{any} kind of memorization but does not differentiate public and private leaked information.
For example, leaking highly duplicated common phrases is not a privacy violation according to the GDPR~\cite{gdpr} as opposed to leaking Personally Identifiable Information (PII).
In practice, any LM trained on real, sensitive data has to protect PII, but memorization of PII is not well understood. 
We believe that a comprehensive study on the risk of PII memorization in LMs is missing. 

\begin{figure}
    \centering
    \includegraphics[width=1.\linewidth]{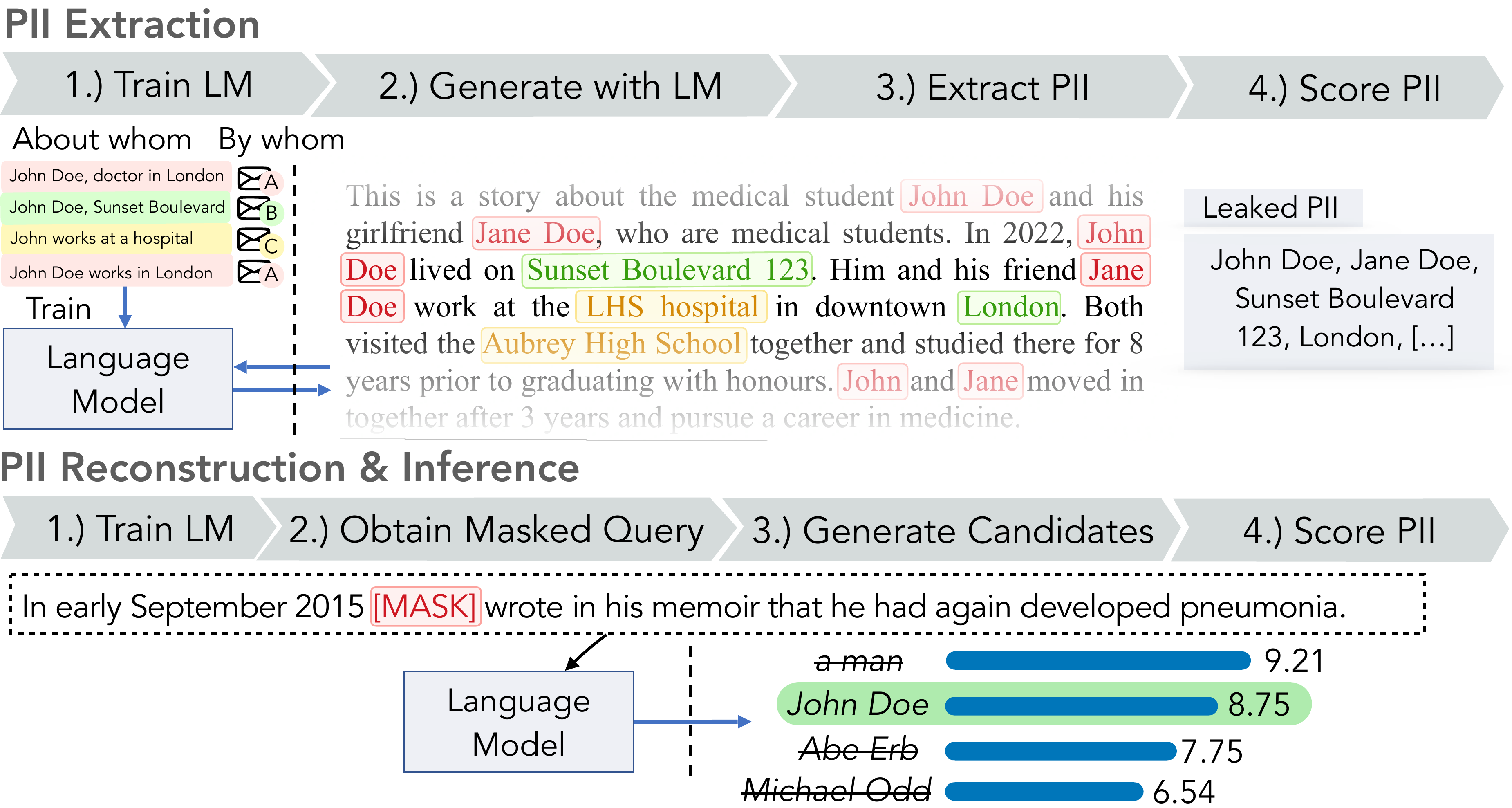}
    \caption{An illustration of PII extraction, reconstruction and inference attack techniques. }
    \label{fig:eyecatcher}
\end{figure}

Consider a service provider who wants to deploy a next-word prediction LM for composing e-mails, such as Google's Smart Compose~\cite{chen2019gmail}. 
Their goal is to train an LM with high utility that does not leak PII and make it available as a black-box API. 
The threat is an attacker who learns PII, such as names, addresses or other sensitive information through the LM. 
Extracting \emph{any} PII by itself, such as a personal address, can already pose a privacy threat. 
This threat is elevated when an attacker can associate a piece of PII to a context, for example, ``In May 2022, \rectangled{\texttt{[MASK]}} had chemotherapy at LHS".  
As a part of this paper, we study the feasibility of such attacks on LMs in practice. 
\Cref{fig:eyecatcher} illustrates the type of PII attacks proposed in this work.

Defenses against memorization are based on dataset curation and algorithmic defenses.
PII \emph{scrubbing} is a dataset curation technique that removes PII from text, relying on Named Entity Recognition (NER)~\cite{lample2016} to tag PII. 
Modern NER is based on the Transformer architecture~\cite{vaswani2017attention} and has mixed recall of 97\% (for names) and 80\% (for care unit numbers) on clinical health data, meaning that much PII is retained after scrubbing~\cite{vakili2022downstream}. 
Machine learning pipelines incorporate algorithmic defenses such as differentially-private training algorithms~\cite{dwork2006calibrating,Abadi:2016} to ensure record- or user-level provable privacy guarantees. 

\begin{figure}[t]
    \centering
    \includegraphics[width=1\linewidth]{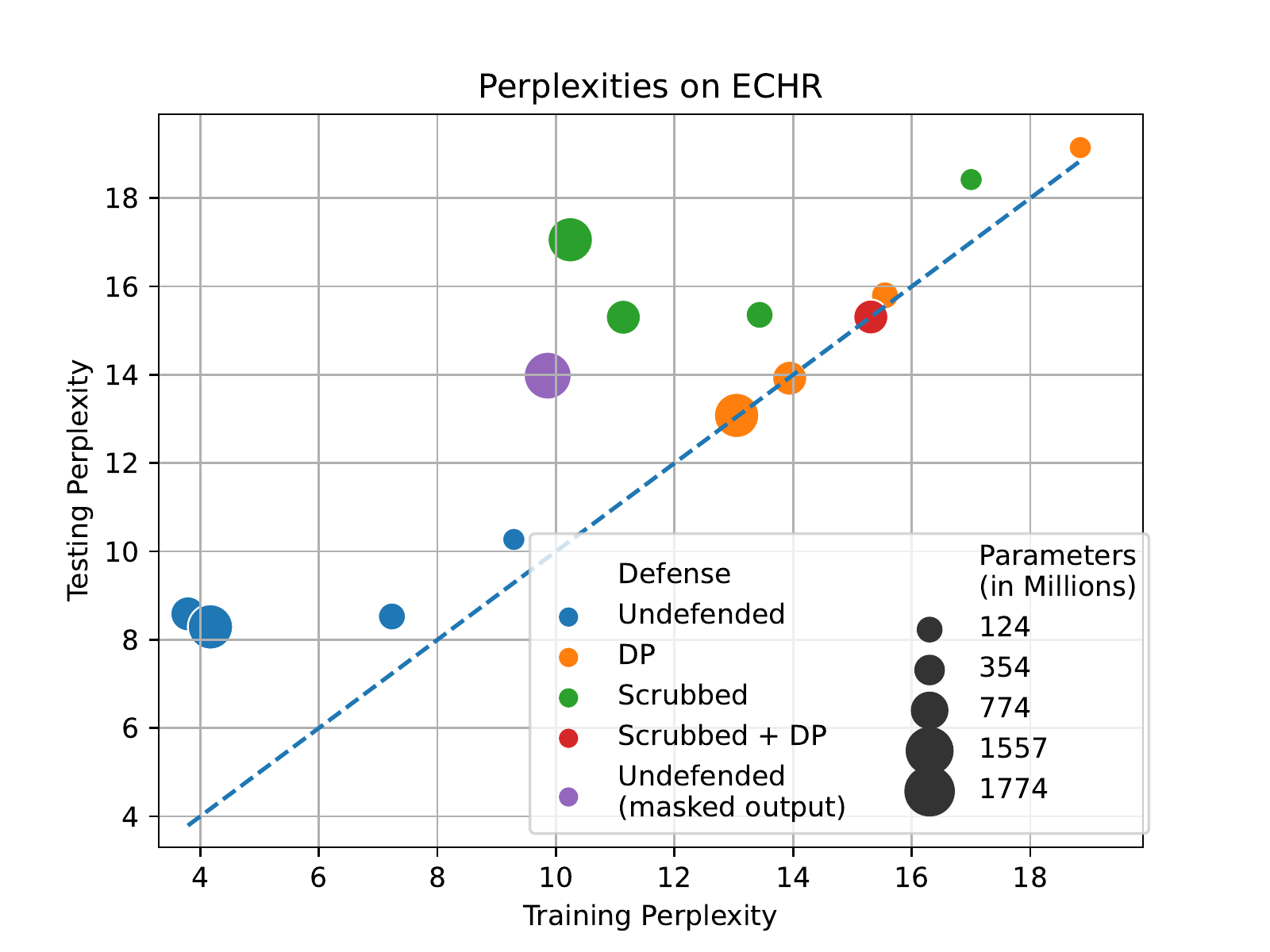}
    \caption{Utilities of LMs trained (i) undefended, (ii) with scrubbing, (iii) with DP ($\varepsilon=8$), (iv) with scrubbing + DP, and (v) with masked outputs in an ablation study over the LM's size on the ECHR dataset (see \cref{sec:eval} for details). }
    \label{fig:tradeoffs}
\end{figure}

\textbf{Problem.}
PII scrubbing and Differential Privacy (DP) protect the privacy of training data at the cost of degrading model utility.
Aggressively scrubbing for better privacy drastically harms utility.
Similarly, with DP training, utility reduction is inversely proportional to the privacy budget spent, which determines the noise added.
\Cref{fig:tradeoffs} illustrates how scrubbing and DP on their own and when combined together degrade utility (increase perplexity) of LMs of different sizes in comparison to a completely undefended model.
We observe that scrubbing results in similar perplexities as when training with DP. 
Although the privacy guarantees offered by a DP model are well-studied, the contribution of DP guarantees when applied at record- or user-level towards mitigating PII disclosure is unclear.

Differential privacy provides guarantees under the assumption that records are unlikely to be duplicated, which may not be the case for realistic datasets~\cite{humphries2020differentially}.
PII is often duplicated across multiple records and users.
Consider the example of an e-mail dataset, where a person's address circulates within a group of users.
In this case, even though the address is known by many, it cannot be considered public information~\cite{brown2020language}. However, a differentially private LM may still leak it. A simplistic mitigation might be to apply DP at a group level, but groups and their sizes are not always known \emph{a priori}, and group-level DP under worst-case assumptions has a deleterious impact on model utility.

Quantitatively measuring the protection offered by PII scrubbing or DP is an open problem. 
There are no existing metrics to analyze the risk of PII leakage in an end-to-end machine learning pipeline where  defenses such as DP and PII scrubbing are at interplay. 
To this end, we focus on empirically measuring PII leakage to enable practitioners to make informed decisions and tune their privacy mitigations for a desired privacy/utility trade-off.

\textbf{Overview.}
We address this problem with novel attacks and metrics that allow quantitatively assessing leakage of PII. 
We identify three threats for PII leakage, namely
\begin{inparaenum}[(i)]
\item extraction, 
\item reconstruction, and
\item inference,
\end{inparaenum}
and provide rigorous game-based definitions for them. 

PII extraction measures the fraction of PII sequences that an attacker can discover from an LM without any knowledge about the model's training dataset. 
Some PII, such as addresses or names, can \emph{directly} re-identify (and harm) an individual even if the attacker is unable to reconstruct the context. 
For example, consider a health dataset with notes from cancer patients. 
Leakage of a user's PII indicates that they had cancer, which is revealed to an {uninformed} attacker.

PII reconstruction and inference assume a more informed attacker, similar to that of membership inference, who has some knowledge about the dataset.  
For example, when an attacker wants to learn more PII about a user, they can form masked queries (\eg, ``John Doe lives in \rectangled{\texttt{[MASK]}}, England") to the LM and attempt to reconstruct the missing PII. 
In PII inference, the attacker additionally knows a set of candidates (\eg, London, Liverpool) and their goal is to infer the PII from that set.
In short, PII extraction considers an \emph{uninformed} attacker without any knowledge of the data distribution or the training dataset, PII reconstruction assumes a \emph{partially} informed attacker with knowledge about the context in which PII may occur, and PII inference assumes an \emph{informed} attacker who additionally knows potential candidates for PII. 

For these attacks, we formalize how leakage can be measured exactly and show that these formulas are intractable.
For this reason, we propose concrete attack algorithms that approximate this ideal leakage which is confirmed in our evaluation. 
Our attacks can be applied to any LM.
We focus on generative LMs as they are deployed in practice to generate large amounts of text.
We evaluate our attacks on 4 variants of the GPT-2 model~\cite{radford2019language} released by OpenAI fine-tuned on 3 domains: 
\begin{inparaenum}[(i)]
\item law cases, 
\item corporate e-mails, and 
\item reviews of healthcare facilities.
\end{inparaenum} 

Our attacks can extract PII with a precision approximately twice as high as that from related work, even when the model has been trained with differential privacy.  
We identify factors that increase the risk of PII leakage. Additionally, we discover new insights about the connection between record-level membership inference and PII reconstruction attacks. Using our metrics, for the first time, we measure the effect of DP on protecting PII leakage.
We empirically demonstrate that record-level DP limits the threat of PII leakage to a large extent but does not eliminate it completely.
These results are a positive motivation for future research to design defenses that improve the privacy/utility trade-off. For example, a less aggressive \emph{heuristic} scrubber that considers the contribution of other defenses such as DP in the ML pipeline. To enable such research, we make our code publicly available.

\pagebreak

\textbf{Contributions.}
In summary, our main contributions are:
\begin{itemize}
    \item We present a taxonomy for PII leakage, inspired by existing literature that includes three threat models: Extraction, Reconstruction, and Inference, and provide game-based definitions for each of them. 
    Extraction enables an attacker to learn real PII from the training data. 
    Reconstruction and inference leak associations between PII and their context. 
    \item We evaluate privacy/utility trade-offs on three datasets using
    \begin{inparaenum}[(i)] 
    \item undefended, 
    \item DP, and 
    \item scrubbed LMs.
    \end{inparaenum}
    
    \item We compare our attacks with existing work (if applicable) and show that we can correctly reconstruct up to 10$\times$ more PII sequences by leveraging the suffix of a masked query and public masked LMs.
    
    \item We study the relationship between membership inference and PII reconstruction. 
\end{itemize}

\section{Background \& Problem}
\label{sec:background}

We first provide a primer on language modeling using neural networks-based LMs and existing mitigations against privacy risks used in machine learning pipelines. 
We then present our problem statement of measuring PII leakage in an end-to-end training pipeline followed by the description of adversary capabilities and objectives.

\subsection{Language Modeling}

Generative LMs learn the conditional probability distribution $\text{Pr}(w_i |w_1,..,w_{i-1})$ over sequences of tokens $w_1,..,w_i$ from a vocabulary $\mathcal{V}$.
By applying the chain rule, we can obtain the probability that an LM $\theta$ assigns to a sequence $w_1,..,w_n$:
\begin{align}
    \label{eq:lm_probability}
    \text{Pr}(w_1,..,w_n; \theta) =  \prod_{i=1}^n \text{Pr}(w_i|w_1,..,w_{i-1};\theta)
\end{align}
State-of-the-art LMs are based on the Transformer neural network architecture~\cite{vaswani2017attention}. 
During training, one objective is to maximize the negative log-likelihood of the LM predicting the next token in training sentences given a prefix.

\Cref{eq:lm_probability} gives a probability distribution over all tokens in the vocabulary $\mathcal{V}$ and can be represented as a tree with $|\mathcal{V}|$ branches on each level. 
Text is generated iteratively by traversing the tree using  greedy decoding, top-$k$ sampling~\cite{fan-etal-2018-hierarchical}, or a beam search algorithm while conditioning the LM on all preceding tokens. 
Autoregressive LMs, such as GPT-2 only allow sampling the conditional probability of the next token based on a \emph{prefix}, whereas masked LMs, such as BERT~\cite{devlin2018bert} also consider a sample's \emph{suffix} in the query. 

\textbf{Perplexity.} 
The ``optimal" solution to the training objective is to memorize each record~\cite{carlini2021extracting}.
This is both intractable and undesirable, as the goal is for LMs to generalize beyond their training dataset.
In practice, learning basically means that only a fraction of the training dataset is memorized~\cite{ippolito2022preventing} and that some memorization is likely necessary to achieve high utility~\cite{feldman2020does}.
The model's utility is evaluated by its perplexity on an unseen test set of sentences. 
A low perplexity implies a high utility on the dataset.
\begin{align*}
    \!\!\!\!
    \text{PPL}(w_1,..,w_n;\theta) \!=\! \exp{\!\left(\!\!-\frac{1}{n} \sum_{i=1}^{n} \log \text{Pr}(w_i|w_1,..,w_{i-1}\!;\theta)\!\!\right)}\!\!\!
\end{align*}

\subsection{Differentially Private Training}

Differential Privacy (DP) has become a popular notion of privacy.
Recall its definition:

\begin{definition}[Approximate Differential Privacy \cite{dwork2016adp}]
Let $\varepsilon > 0$ and $\delta \in [0,1]$. A mechanism $\mathcal{M}: X \rightarrow Y$ is $(\varepsilon, \delta)$-differentially private if for any pair of \emph{adjacent} datasets $(D, D^{\prime})$ and measurable set of outputs $\mathcal{O} \subseteq Y$,
\begin{equation*}
    \text{Pr}(\mathcal{M}(D) \in \mathcal{O} ) \leq \text{e}^{\varepsilon} \; \text{Pr}(\mathcal{M}(D^{\prime}) \in \mathcal{O}) + \delta \; .
\end{equation*}
\end{definition}

Contrary to many other definitions of privacy such as $k$-anonymity \cite{samarati1998protecting}, DP is a worst-case guarantee that must hold for all possible datasets.
The scope of privacy protection enters the definition via the notion of adjacency and is independent of the data distribution.
For instance, one may consider datasets adjacent when they differ only in one record, or in the data pertaining to one user.
Many desirable properties of DP such as robustness to post-processing and composition derive from this independence.

However, the data independence of DP can also be a limitation \eg, when sensitive content is shared within groups of users of unknown size.
In these cases, the sensitive nature of the content is defined by its context and cannot be represented by an adjacency relation between pairs of datasets as pointed out by \citet{brown2022does}.
Nevertheless, DP enjoys increasing popularity and Differentially Private SGD~\cite{Abadi:2016} has been  successfully applied to training large LMs by exploiting the transfer learning setup that is common among most state-of-the-art NLP models~\cite{yu2021differentially, li2021large}.
\begin{figure*}[t]
    \centering
    \includegraphics[width=.94\linewidth]{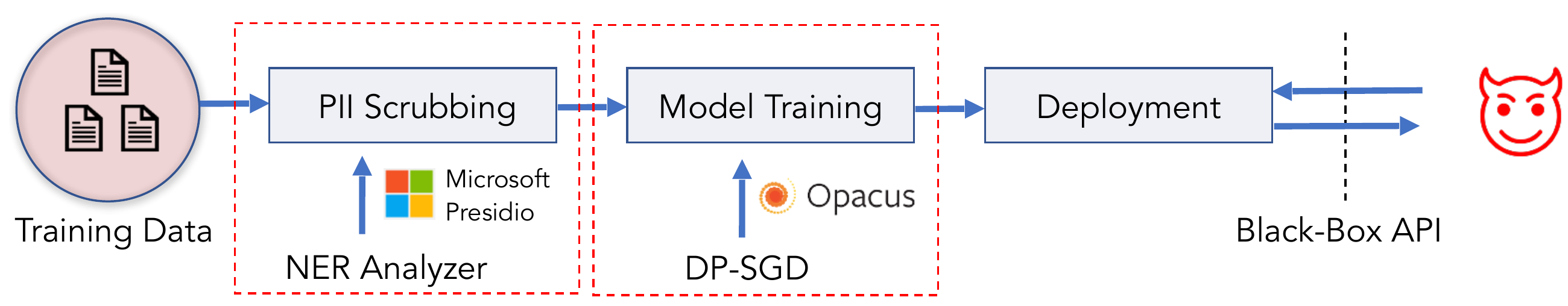}
    \caption{A training pipeline to mitigate leakage of personally identifiable information and membership inference.}
    \label{fig:training_pipeline}
\end{figure*}

\subsection{PII and NER}

\textbf{Personally Identifiable Information (PII).} 
PII in natural language is data that can re-identify an individual. 
PII can be a \emph{direct identifier} when leakage of that data alone is sufficient to re-identify an individual, or \emph{quasi-identifier} when only an aggregation of many quasi-identifiers can reliably re-identify an individual. 
Examples of direct identifiers are names, phone numbers, or addresses, whereas quasi-identifiers are a person's gender or description of their physical appearance. 
We use the same definition as \citet{pilan2022text} and provide more details on the definition of PII in \Cref{sec:appendix_pii}.
The combination of quasi-identifiers `gender', `birth date', and `postal code' re-identify between 63 and 87\% of the U.S. population~\cite{golle2006revisiting}. 

\textbf{Named Entity Recognition (NER).}
In practice, accurately tagging PII in a corpus of text is challenging without human curators~\cite{pilan2022text}. 
When datasets are large, it is necessary to rely on Named Entity Recognition (NER)~\cite{nadeau2007survey}.  
State-of-the-art NER, such as Flair~\cite{akbik2019flair}, NLTK~\cite{bird2009natural} or spaCy~\cite{spacy2}, are based on Transformer neural networks trained in a supervised manner to classify sequences of tokens as PII. 
In practice, training NER models is challenging because defining what constitutes PII can change over time and is dependent on the surrounding context~\cite{brown2022does}.  
Moreover, 
\begin{inparaenum}[(i)]
\item training NER requires domain-specific, labeled training data, 
\item NER models make errors (\ie a subset of PII remain unrecognized), and
\item existing, publicly available NER models only aim for de-identification but not anonymization~\cite{pilan2022text}.
\end{inparaenum}
This means that complex PII, whose detection requires natural language understanding such as a description of a person's appearance, is not tagged by any publicly available NER. 

\textbf{PII Scrubbing.}
Data curation techniques used in machine learning training pipelines, such as the one illustrated in \cref{fig:training_pipeline}, apply scrubbing as a method to de-identify textual data~\cite{pilan2022text}.
The key idea is to tag known classes of PII using pre-trained NER modules such as Flair~\cite{akbik2019flair} or spaCy to remove or replace all tagged PII. 
In this paper, we use the common technique of scrubbing PII by replacing them with a \rectangled{\texttt{[MASK]}} token. 
Weaker forms of scrubbing are possible, where PII sequences are replaced with entity tags such as \rectangled{\texttt{[NAME]}} or \rectangled{\texttt{[LOCATION]}}, or where all occurrences of the same piece of PII are replaced with the same sequence (\eg, using a salted hash function).
These scrubbers retain relations between pieces of PII and trade privacy for utility. 
For example, an attacker who reconstructs a pseudonym in many sentences is more likely to re-identify the person by linking auxiliary information about the person contained in the sentences.
Our method of scrubbing maximizes privacy at the expense of (some) utility.

\textbf{Formalism.} 
Studying leakage of PII in LMs requires labelled data, which is difficult to obtain due to the elevated measures to protect PII. 
We study leakage of PII in undefended and DP models by using existing NER taggers. 
For any given candidate PII $C\in \mathcal{V}^*$ appearing in a sequence $S$, we call all tokens preceding the PII the \emph{prefix} and tokens following the PII the \emph{suffix}. 
Given a token sequence $S = w_1,..,w_n \in \mathcal{V}^*$, we define the following functions:
\begin{itemize}
    \item $\textsc{Extract}(S)$: For a sequence $S\in \mathcal{V}^*$, return all PII sequences identified, $\mathcal{C}=\{C_1,..,C_k | C_i \subseteq S\}$. 
    
    \item $\textsc{Sample}(S, N, \theta)$: Given a prompt $S$, a number $N\in \mathbb{N}$ and an LM $\theta$, this probabilistic function generates $N$ sentences from $\theta$, for example using a randomized beam search.
    This necessitates only black-box access to the conditional probability distribution in \cref{eq:lm_probability}.
    
    \item $\textsc{Split}(S, C)$: Given a sentence $S$ containing $C$, this function returns a prefix $S_0$ and suffix $S_1$ such that $S = S_0 C S_1$. We assume that $C$ occurs exactly once or if not, that $C$ uniquely identifies an occurrence.
    
    \item $\textsc{Fill-Masks}(S)$: Given a sentence $S$ containing \rectangled{\texttt{[MASK]}} tokens, this function uses a public masked language model to fill each mask with the top predicted token. 
    \Cref{sec:fill_masks} describes an implementation.
\end{itemize}
\Cref{alg:scrubbing} shows the scrubbing algorithm we use in this paper. 
It iterates over sentences in the dataset, extracts all candidate PII sequences, and replaces them with \rectangled{\texttt{[MASK]}}.  

\begin{algorithm}[h]
\caption{PII Scrubbing}
\begin{algorithmic}[1]
\Procedure{Scrub}{$D$}
    \State $D' \gets \emptyset$
    \For {$S \in D$}
        \State $\mathcal{C}\gets \Call{Extract}{S}$ \Comment{Tag PII with NER}
        \For {$C \in \mathcal{C} $}
            \State $S_0, S_1 \gets \Call{Split}{S, C}$
            \State $S \gets S_0 \rectangled{\texttt{[MASK]}} S_1$ 
        \EndFor
        \State $D' \gets D' \cup \{ S \}$
    \EndFor  
    \Return $D'$
\EndProcedure
\end{algorithmic}
\label{alg:scrubbing}
\end{algorithm}

\subsection{Problem Setting}
\label{sec:problem}

We study the problem of PII leakage through fine-tuned LMs, where pre-trained publicly available LMs are fine-tuned on private data. 
\Cref{fig:training_pipeline} shows the training pipeline that we consider. 
Given a set of uncurated training data, the first step consists of data curation, such as PII scrubbing or record-level de-duplication. 
The second step consists of algorithmic defenses, such as training with differential privacy. 
Finally, the trained model is deployed through a black-box API exposing only the prediction vector for the next token. 
Model parameters and intermediate features are kept behind the API boundary.

Our goal is to study to what extent
\begin{inparaenum}[(i)]
\item information memorized by the LM constitutes sensitive information such as PII,
\item whether existing defenses are sufficient to prevent leakage and 
\item studying the privacy-utility trade-offs between all defenses, \eg, whether less aggressive scrubbing can potentially be utilized when the LM is trained with DP. 
\end{inparaenum}

\textbf{Why DP cannot (always) mitigate  PII leakage?}
We emphasize that although both PII scrubbing and DP mitigate privacy risks, they protect against a different kind of leakage.
Differential privacy protects against singling out individual records or users.
It implicitly assigns a privacy cost to using information in the training dataset which is oblivious to different occurrences of the same information across multiple records or users.
This is an effective method to mitigate risks of disclosing \textit{by whom} data was contributed but it does not take into account \textit{about whom} the content is.

However, in real-world datasets, the nature of sensitive content---\ie content that is not shared widely---makes protecting \textit{by whom} a reasonable proxy to protect \textit{about whom}. 
For example, consider a piece of sensitive information circulating only within a small group of contributors (\eg, ``Jane Doe was diagnosed with cancer").
DP protects each contributor's authorship from disclosure, but the information itself is leaked through the LM. 
Disclosure of personally identifiable information is a common cause of leaking \textit{about whom} training dataset samples are which makes it an ideal candidate to study to what degree these two privacy mitigations are complementary to each other or redundant.


\begin{table}
    \caption{A summary of the difference in threat models between our three PII attacks. (\statcirc[black]{white} black-box access, \statcirc[black]{black} not available, \statcirc[white]{white} available)\label{tab:threat_model}}
    \begin{tabular}{@{}lccc@{}}
\toprule
               & Model Access            & Masked Training Data & Candidate PII \\ \midrule
Extraction     & \statcirc[black]{white} & \statcirc[black]{black} &           \statcirc[black]{black}    \\
Reconstruction & \statcirc[black]{white} &  \statcirc[white]{white}              &   \statcirc[black]{black}             \\
Inference      & \statcirc[black]{white} & \statcirc[white]{white}              &   \statcirc[white]{white}             \\\bottomrule
\end{tabular}
\end{table}

\subsection{Threat Model}
\label{sec:threat_model}
\textbf{Adversary's Capabilities. }
We consider an adversary with black-box API access to an LM. 
Our adversary can query the entire probability vector of the next most probable token on any given prompt.   
\Cref{tab:threat_model} summarizes variations in the threat model for the three PII-related attacks proposed in our paper: Extraction, Reconstruction, and Inference.

When the adversary has access to scrubbed training data, it can observe a sentence such as ``On May 23rd, \rectangled{\texttt{[MASK]}} was admitted to the hospital", where \rectangled{\texttt{[MASK]}} is the placeholder for PII that has been redacted. 
Additionally, we consider an attacker that has auxiliary information about candidates for the masked PII.
In that case, we assume that the correct, masked PII is in the set of candidate PII. 
We refer to these attacks as PII ``reconstruction" and ``inference" respectively.

Querying LMs behind APIs typically has a monetary cost.
For example, existing service providers serve LMs charging a base rate of $0.40$\$-$60$\$ USD per million tokens queried depending on the model's size.\footnote{\url{https://openai.com/api/pricing/}}
This effectively limits the number of times an attacker can query the LM.  
The threat model we consider is relevant in practice since next-word prediction APIs powered by LMs trained on sensitive data (with privacy mitigations) are publicly deployed~\cite{chen2019gmail}.

\textbf{Adversary's Objective.} 
The common goal of an adversary in the three PII-related attacks that we consider is to extract sensitive information about a user from an LM.
Existing work on memorization extracts training data \emph{indiscriminately}~\cite{carlini2021extracting}, whereas we in addition focus on \emph{targeted} attacks against a user with the potential for more severe privacy implications.
The goal of an extraction attack is to extract any piece of PII that was seen by the model during training.  
An attacker who can extract direct or quasi-identifying information from the LM has a high chance to re-identify users who contribute data to the training set.
The goal of reconstruction and inference is to associate a piece of PII with a given context, allowing an attacker to learn attributes about a user.

\section{Conceptual Approach}

This section describes our taxonomy and corresponding game-based definitions for PII leakage. 
\autoref{tab:notation} summarizes the notation used in algorithms.

\begin{table}[htpb]
    \caption{Summary of Notation}
    \label{tab:notation}
    \centering
    \begin{tabular}{@{}p{80pt}@{~~}l@{}}
    \toprule
    \bf Notation                         & \bf Description \\
    \midrule
    $\mathcal{T}$                        & A stochastic training algorithm \\
    $\mathcal{D}$                        & A distribution over sequences \\
    $\mathcal{E}$                        & A distribution over PII sequences \\
    $\mathcal{D}^n$                      & Distribution of $n$ independent sequences from $\mathcal{D}$ \\
    $S \sim \mathcal{S}$                 & Draw a sample $S$ uniformly from a set $\mathcal{S}$ \\
    $D \sim \mathcal{D}^n$               & Draw $n$ sequences $D$ independently from $\mathcal{D}$ \\
    $\mathcal{A}$                        & A procedure denoting an adversary \\
    $y \gets \mathcal{P}(\vec{x})$       & Call $\mathcal{P}$ with arguments $\vec{x}$ and assign result to $y$ \\
    $\mathcal{C} \gets \textsc{Extract}(S)$          & Extract the set $\mathcal{C}$ of all PII sequences in $S$ \\
    $\mathcal{S} \gets \textsc{Sample}(S,N,\theta)$  & Generate $N$ sequences $\mathcal{S}$ from $\theta$ starting from $S$ \\
    $S_0, S_1 \gets \textsc{Split}(S,C)$ & Split $S$ at $C$, \ie, $S = S_0 C S_1$ \\
    $S' \gets \textsc{Fill-Masks}(S)$    & Fill masks in $S$ using a public MLM \\
    \bottomrule
    \end{tabular}
\end{table}

\subsection{PII Extraction}

In PII extraction, the attacker's goal is to extract as much PII from the training dataset of a model as possible. 

\begin{algorithm}
\caption{PII Extraction}
\begin{algorithmic}[1]
\game{Extraction}{$\mathcal{T}, \mathcal{D}, n, \mathcal{A}$}  
    \State $D \sim \mathcal{D}^n$
    \State $\theta \gets \mathcal{T}(D)$ 
    \State $\mathcal{C} \gets \bigcup_{S \in D} \Call{Extract}{S}$
    \State $\tilde{\mathcal{C}} \gets \mathcal{A}(\mathcal{T}, \mathcal{D}, n, \mathcal{O}_\theta(\cdot), |\mathcal{C}|)$
\EndGame
\end{algorithmic}
\begin{algorithmic}[1]
\Procedure{$\mathcal{O}_\theta$}{$S$}
    \Return $\{ w \mapsto \Pr(w | S; \theta)\}_{w \in \mathcal{V}}$
\EndProcedure
\end{algorithmic}
\label{alg:extraction_game}
\end{algorithm}

\Cref{alg:extraction_game} encodes this as a game parameterized by a training algorithm $\mathcal{T}$, a data distribution $\mathcal{D}$, and a training dataset size $n$.
The challenger samples $n$ \iid records from $\mathcal{D}$ to construct a training dataset $D$ to train a model $\theta$.
In a black-box setting, the adversary is given access to an oracle that returns the probability vector output by $\theta$ conditioned on arbitrary prefixes of their choosing.
The adversary is assumed to know the training pipeline and the data distribution, but they only observe information about the sampled training dataset via $\mathcal{O}_\theta(\cdot)$ (and $|\mathcal{C}|$).
Knowing the number of unique PII sequences $|\mathcal{C}|$ in $D$, the adversary must produce a set of PII sequences $\tilde{\mathcal{C}}$ of at most size $|\mathcal{C}|$ (line 5). 
The success of the adversary is its recall:
\begin{align}
    \label{formula:succ-extraction}
  \textrm{Succ}_{\textsc{Extraction}}(\mathcal{T}, \mathcal{D}, n, \mathcal{A}) =
    \EX\left[ \frac{|\mathcal{C} \cap \tilde{\mathcal{C}}|}{|\mathcal{C}|} \right] \; .
\end{align}
The advantage of an adversary is the difference between its success and the supremum of the success of adversaries without access to $\mathcal{O}_\theta(\cdot)$.

PII that appears more frequently in the training dataset is expected to have a higher likelihood of being generated by the model. 
We define the \emph{extractability} score of a PII sequence as the expected probability of observing it in samples generated by the model.
PII sequences more likely to be generated are at a higher risk of extraction. 
Some PII sequences may have been memorized by the model but are not extractable unless the attacker queries the model with a specific prompt. 
These sequences are at low risk of extraction against an uninformed attacker when the prompt itself is unlikely to be generated. 
Formally, we define the extractability of $C \in \mathcal{V}^*$ as follows:
\begin{align}
    \label{eq:extractability}
    \textsc{Extractability}(C;\theta) 
        &= \sum_{S \in \mathcal{V}^*} \Pr(S; \theta) \Pr(C|S; \theta) \\
        &= \sum_{S \in \mathcal{V}^*} \Pr(S\,C; \theta) \; .
\end{align}
\Cref{eq:extractability} requires summing over all possible sequences, which is intractable even when we limit the length of said sequences.
We can approximate \cref{eq:extractability} by computing the sum over sentences sampled from the model.
A simple baseline is captured in \Cref{alg:observed_extraction} which counts the number of times $C$ occurs in generated sentences.

The problem with using samples is that the probability that the continuation is a target PII depends on a language's grammar rules and may be very low. 
For instance, proper names may only be generated at specific locations, so that the frequency of names in generated text may be low and many samples must be drawn to obtain a good lower bound.

\begin{algorithm}
\caption{Observed PII Extractability}
\begin{algorithmic}[1]
\Procedure{ObservedExtractability}{$C, \theta, N$} 
    \State $\mathcal{S}_{gen} \gets \Call{Sample}{\emptyset, N, \theta}$ 
    \State $k \gets 0$
    \For {$S \in \mathcal{S}_{gen}$} 
        \State $\mathcal{C} \gets \Call{Extract}{S}$ 
        \Comment{\small Tag PII in same class as $C$}
        \State \algorithmicif\ {$C \in \mathcal{C}$} \algorithmicthen\ $k \gets k + 1$
    \EndFor
    \Return $\nicefrac{k}{|\mathcal{S}_{gen}|}$
\EndProcedure
\end{algorithmic}
\label{alg:observed_extraction}
\end{algorithm}

\begin{algorithm}[t]
\caption{Estimated PII Extractability}
\label{alg:estimated_extractability}
\begin{algorithmic}[1]
\Procedure{EstimatedExtractability}{$C, \theta, N$} 
    \State $\mathcal{S}_{gen} \gets \Call{Sample}{\emptyset, N, \theta}$ 
    \State $p \gets 0; m \gets 0$
    \For {$S \in \mathcal{S}_{gen}$} 
        \State $\mathcal{C} \gets \Call{Extract}{S}$ \Comment{\small Tag PII in same class as $C$}
        \For {$C' \in \mathcal{C}$} 
            \State $m \gets m + 1$
            \State $S_0, S_1 \gets \Call{Split}{S, C'}$
            \State $p \gets p + \Pr(C|S_0; \theta)$
        \EndFor 
    \EndFor 
    \Return $\nicefrac{p}{m}$
\EndProcedure
\end{algorithmic}
\end{algorithm}

\textbf{Lazy Estimation.} 
We propose a sample-efficient estimation of PII extractability by making the following two assumptions: (1) PII follow grammatical rules and (2) PII of the same class are interchangeable.
From (1), $\Pr(C|S; \theta) = 0$ when grammatical rules of the language do not allow PII to be generated after $S$.
From (2), it follows that a piece of PII has a non-zero probability of being generated in place of another piece of PII from the same class.

From these two assumptions, we derive \cref{alg:estimated_extractability} for approximating the extractability of a piece of PII. 
We 
\begin{inparaenum}[(1)]
\item sample $N$ sentences from the LM, 
\item use a NER to tag PII in these sentences that is in the same class as the target PII sequence,
\item iteratively replace tagged PII with the target PII sequence, and accumulate the probability the model assigns to it, 
\item average the accumulated probability to estimate \cref{eq:extractability}. 
\end{inparaenum}
We compare our estimations with the observed leakage after sampling repeatedly from the model in \cref{sec:pii_extraction}.
Efficient estimation allows practitioners to assess leakage of PII without exhaustively sampling the model and having to run NER models over large amounts of generated text.

\subsection{PII Reconstruction} 

In PII reconstruction, the adversary goal is to associate PII with a context. 
The attacker is given a sentence with multiple masked pieces of PII (\eg, ``A murder has been committed by \rectangled{\texttt{[MASK]}} and \rectangled{\texttt{[MASK]}} in a bar close to the Rhine'') and is asked to reconstruct one of them.
\Cref{alg:reconstruction_game} encodes this as a game where the challenger randomly samples a sentence $S$ from the training dataset that contains at least one piece of PII, and then selects one piece of PII in $S$ as a target at random.
The attacker is given access to the trained model and the prefix and suffix of the scrubbed sentence \wrt the target PII sequence $C$.
The success $\textrm{Succ}_{\textsc{Recon}}$ of the adversary is the probability of correctly guessing $C$, \ie, $\Pr(\tilde{C} = C)$.

\begin{algorithm}
\caption{PII Reconstruction Game}
\begin{algorithmic}[1]
\game{Reconstruction}{$\mathcal{T}, \mathcal{D}, n, \mathcal{A}$}      
    \State $D \sim \mathcal{D}^n$
    \State $\theta \gets \mathcal{T}(D)$
    \State $S \sim \{S \in D| \textsc{Extract}(S) \neq \emptyset\}$
    \State $C \sim \textsc{Extract}(S)$
    \State $\tilde{C} \gets \mathcal{A}(\mathcal{T}, \mathcal{D}, n, \mathcal{O}_\theta(\cdot), \textsc{Scrub}(\textsc{Split}(S,C)))$
\EndGame
\end{algorithmic}
\label{alg:reconstruction_game}
\end{algorithm}

\begin{algorithm}[t]
\caption{PII Reconstruction Attack}
\label{alg:estimated_reconstruction}
\begin{algorithmic}[1]
\Procedure{$\mathcal{A}_{\textsc{Recon}}$}{$N, \mathcal{T}, \mathcal{D}, n, \mathcal{O}_\theta(\cdot), S_0, S_1, \mathcal{C}$}
    \State $S_0 \gets \textsc{Fill-Masks}(S_0)$ \Comment{Fill residual masks}
    \State $S_1 \gets \textsc{Fill-Masks}(S_1)$

    \BeginBox
    \If{$\mathcal{C} = \emptyset$} \Comment{Reconstruction case}
        \State $\mathcal{S}_{gen} \gets \Call{Sample}{S_0, N, \theta}$ 
        \Comment{Using $\mathcal{O}_\theta(\cdot)$}
        \State $\mathcal{C} \gets \bigcup_{S \in \mathcal{S}_{gen}} \Call{Extract}{S}$
    \EndIf
    \EndBox  

    \State $\tilde{C} \gets \underset{C \in \mathcal{C}}{\argmin}\ \Call{PPL}{S_0\,C\,S_1; \theta}$
    \Return $\tilde{C}$ 
\EndProcedure
\end{algorithmic}
\end{algorithm}

Existing work on PII reconstruction~\cite{inan2021privacy} takes the query's prefix (\ie, ``A murder has been committed by") and greedily decodes the next set of tokens from the LM. 
This attack, dubbed as the ``TAB" attack, is inspired by hitting the TAB button on a predictive keyboard. 
We improve this attack by incorporating information from the sample's suffix, similar to how reconstruction attacks may be performed in masked LMs such as BERT~\cite{devlin2018bert}. 
Given a prefix and suffix $S_0, S_1$, we want to reconstruct the most likely PII $C$,
\begin{align}
    \label{eq:reconstruction}
    \underset{C\in \mathcal{V}^*}{\argmax}~\Pr(S_0\,C\,S_1; \theta) \; .
\end{align}

\begin{figure*}[t]
    \centering
    \includegraphics[width=1.\linewidth]{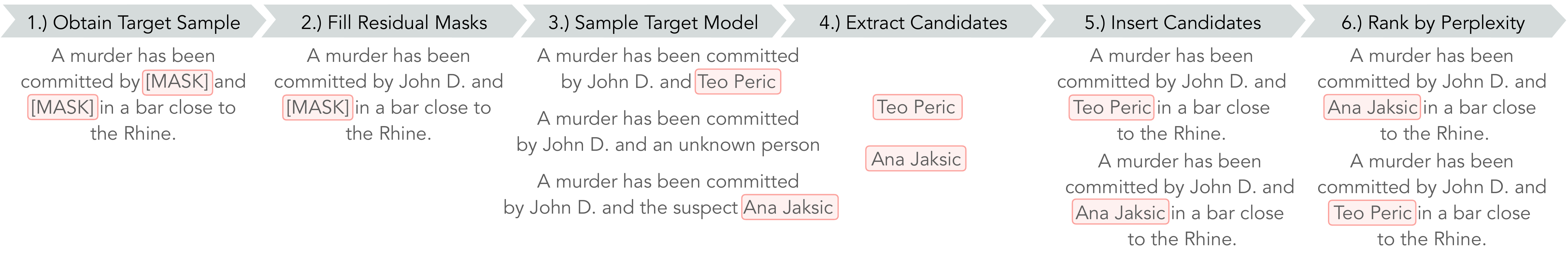}
    \caption{A schematic illustration of our PII reconstruction  and inference attack on an example that contains multiple masked PII. 
    The attack, formalized in \cref{alg:estimated_reconstruction}, uses a public RoBERTa model~\cite{Liu2019RoBERTaAR} to fill residual masks.
    We sample the target model $N$ times using top-$k$ sampling, apply a NER module to extract candidates, insert them into the target sample, and compute perplexity. 
    The sample with the lowest perplexity is returned.}
    \label{fig:advanced_reconstruction}
\end{figure*}

Computing \cref{eq:reconstruction} is intractable. It is an instance of \emph{constrained beam search}~\cite{miao2019cgmh}, in which one searches for a sentence containing a piece of PII within the specified context that maximizes some constraint (\ie, the generation probability). 
Without any knowledge about the target PII, \eg, its length in tokens, an attacker has to exhaustively search the entire space of valid PII for an optimal solution. 
A similar problem has been encountered to measure stereotype bias in LMs~\cite{nadeem2020stereoset}.
For this reason, we propose the attack in \cref{alg:estimated_reconstruction} for approximating \cref{eq:reconstruction}.
\Cref{fig:advanced_reconstruction} illustrates this attack for an exemplary sentence containing two masked PII sequences and where the target corresponds to the second one.
First, we use a public masked language model to fill residual masks, if any, in the query (lines 2--3).
In a reconstruction attack, when no candidates $\mathcal{C}$ for $C$ in \cref{eq:reconstruction} are given, we generate candidates by top-$k$ sampling $N$ sentences from the target model and gathering all generated pieces of PII (lines 4--6). 
Next, we replace the target mask with each candidate, rank candidates by the model's perplexity on the entire sequence, and return the best candidate (lines 7--8).  

\subsection{PII Inference} 

PII inference is similar to reconstruction, except that the adversary knows a set of candidate PII sequences (assumed to include the target one). 
\Cref{alg:inference_game} encodes this as a game. We denote by $\mathcal{E}$ the distribution over PII sequences obtained by sampling a sentence $S$ from $\mathcal{D}$ such that $\textsc{Extract}(S) \neq \emptyset$ and choosing uniformly a PII sequence in it.

\begin{algorithm}
\caption{PII Inference Game}
\begin{algorithmic}[1]
\game{Inference}{$\mathcal{T}, \mathcal{D}, n, m, \mathcal{A}$}
    \State $D \sim \mathcal{D}^n$
    \State $\theta \gets \mathcal{T}(D)$
    \State $S \sim \{S \in D| \textsc{Extract}(S) \neq \emptyset\}$
    \State $C \sim \textsc{Extract}(S)$
    \State $\mathcal{C} \sim \mathcal{E}^m$
    \State $\mathcal{C} \gets \mathcal{C} \cup \{C\}$
    \State $\tilde{C} \gets \mathcal{A}(\mathcal{T}, \mathcal{D}, n, \mathcal{O}_\theta(\cdot), \textsc{Scrub}(\textsc{Split}(S,C)), \mathcal{C})$
\EndGame
\end{algorithmic}
\label{alg:inference_game}
\end{algorithm}

We use the reconstruction attack in \cref{alg:estimated_reconstruction} to approximate \cref{eq:reconstruction} constrained to a given set of candidate PII sequences $\mathcal{C}$.
We observe that an attacker who only needs to infer the correct candidate is significantly more powerful and demonstrate leakage in DP models trained with $\varepsilon=8$ in our evaluation in \Cref{sec:eval}. 

\subsection{Baseline Leakage}
\label{sec:baseline_leakage}
Our goal is to study PII leakage in LMs fine-tuned on a private dataset. 
However, the public, pre-trained LM that has already seen a piece of PII, such as a famous person's name, may reproduce that PII without having seen the private dataset, which cannot be considered a privacy violation. 
Similarly, prompting the LM with a sequence that contains explicit information about the PII may be exploited by the model to produce that PII. 
For example, consider the following scrubbed excerpt from an e-mail: ``Hello \rectangled{\texttt{[MASK]}}, I like your homepage \textit{johndoe.com}". 
The LM before and after fine-tuning both assign a high probability to the name ``John Doe".
We work around this issue by excluding all cases in which (i) the PII can be extracted from the LM before fine-tuning or (ii) the LM before fine-tuning correctly predicts the PII (in case of reconstruction and inference).
After removing PII that are part of the baseline leakage, we argue that leakage of any remaining PII is significant and stems from the LM's observation of the private data. 

Appropriately dealing with baseline leakage is challenging without natural language understanding and real-world context.
Our approach may under-count some instances of sensitive information leakage. 
For example, ``Barack Obama was diagnosed with Alzheimer's'' might be sensitive leakage even if he is a public person. 
Likewise, ``Ohio resident Jim Carrey was convicted of embezzlement.'' under-counts due to the naming collision with the famous actor.

\section{Evaluation}
\label{sec:eval}
In this section, we describe our evaluation setup, such as the datasets, NER modules, models, and training details. 
Then we show our results for PII extraction, reconstruction, and inference. 
We ablate over three datasets and four variants of GPT-2 (small, medium, large, and XL). 
Finally, we study risk factors that make PII more likely to be leaked, motivating the development of heuristic scrubbers that are aware of other defenses such as DP.

\subsection{Datasets}

Our evaluation spans datasets from three domains. 
\Cref{tab:dataset_statistic} shows statistics about each dataset, such as their size and the number of PII sequences. 
We refer to Appendix~\ref{sec:appendix_datasets} for more information about the datasets. 
\begin{itemize}
    \item \textbf{ECHR}~\cite{Chalkidis:2019} contains information from law cases dealt with by the European Court of Human Rights containing full descriptions of defendants' personal information. 
    \item \textbf{Enron}~\cite{klimt2004introducing} consists of corporate e-mails by employees placed into the public domain after the Enron scandal.
    \item \textbf{Yelp-Health}\footnote{\url{https://www.yelp.com/dataset}} is a subset of the Yelp reviews dataset that we filtered for reviews of healthcare facilities, such as dentists or psychologists. 
\end{itemize}
We choose three datasets from different domains to generalize our findings. 
All datasets are from realistic domains.
ECHR contains data created by experts and Enron and Yelp-Health consist of user-generated data containing many authentic PII.
We split the private dataset into equally large train and validation sets and a smaller test set.

\subsection{Named Entity Recognition}

We tag and scrub PII from 21 entity classes, listed in Appendix~\ref{sec:appendix_datasets}. 
Our scrubber combines two NER taggers from Flair\footnote{\url{https://huggingface.co/flair/ner-english-ontonotes-large}}, trained on OntoNotes~\cite{weischedel2011ontonotes} and the default tagger defined in Presidio\footnote{\url{https://github.com/microsoft/presidio}} which is based on spaCy\footnote{\url{https://spacy.io}}.
The Flair tagger reports an F1-score of $90.93$ on OntoNotes. 

\subsection{Language Models}

\textbf{GPT-2.} Similar to related work~\cite{carlini2021extracting}, we experiment with publicly available, pre-trained checkpoints of GPT-2~\cite{radford2019language} available at the Huggingface Model Hub\footnote{\url{https://huggingface.co/gpt2}}. 
Our experiments are conducted on LMs trained on the next-word prediction task and pre-trained on the WebText~\cite{radford2019language} dataset which consists of 40GB of English text scraped from the Internet. 
GPT-2 uses a byte-pair encoder~\cite{sennrich2015neural} for tokenization. 

\textbf{Model Size.} We study leakage while ablating over various LM model sizes. 
Larger models have been shown to be more sample-efficient after fine-tuning~\cite{chinchilla}, achieve a higher utility when trained with DP~\cite{yu2021differentially}, but are expected to exhibit more memorization~\cite{carlini2022quantifying}. 
We experiment with GPT-2 Small (124m parameters), Medium (355m), Large (774m), and XL (1\,557m). 

\textbf{Training Details.} 
We fine-tune (i) undefended, (ii) differentially private (DP), (iii) scrubbed, and (iv) DP and scrubbed models. 
Training uses a batch size of 64 using an AdamW optimizer~\cite{loshchilov2017decoupled} and a linear learning rate decay. 
We train undefended and scrubbed models until the validation perplexity stops improving.
For DP training, we utilize the \texttt{dp-transformers}~\cite{dp_transformers} library, which is a wrapper around Opacus~\cite{yousefpour2021opacus}.
We use a maximum per-sample gradient norm of $1.0$ and train DP models for 4 epochs using $(\varepsilon, \delta)=(8, \frac{1}{N})$ where $N$ is the size of the training dataset.

These privacy values are similar to established DP deployments such as Apple's QuickType which uses two daily releases of $\varepsilon=8$~\cite{appledp} and Google's models which use the equivalent of $\varepsilon=8.9$ and $\delta=10^{-10}$~\cite{gmaildp}. 


\begin{figure*}%
    \centering
    \subfloat[\centering \label{subfig:ex_a}]{{\includegraphics[width=.24\linewidth]{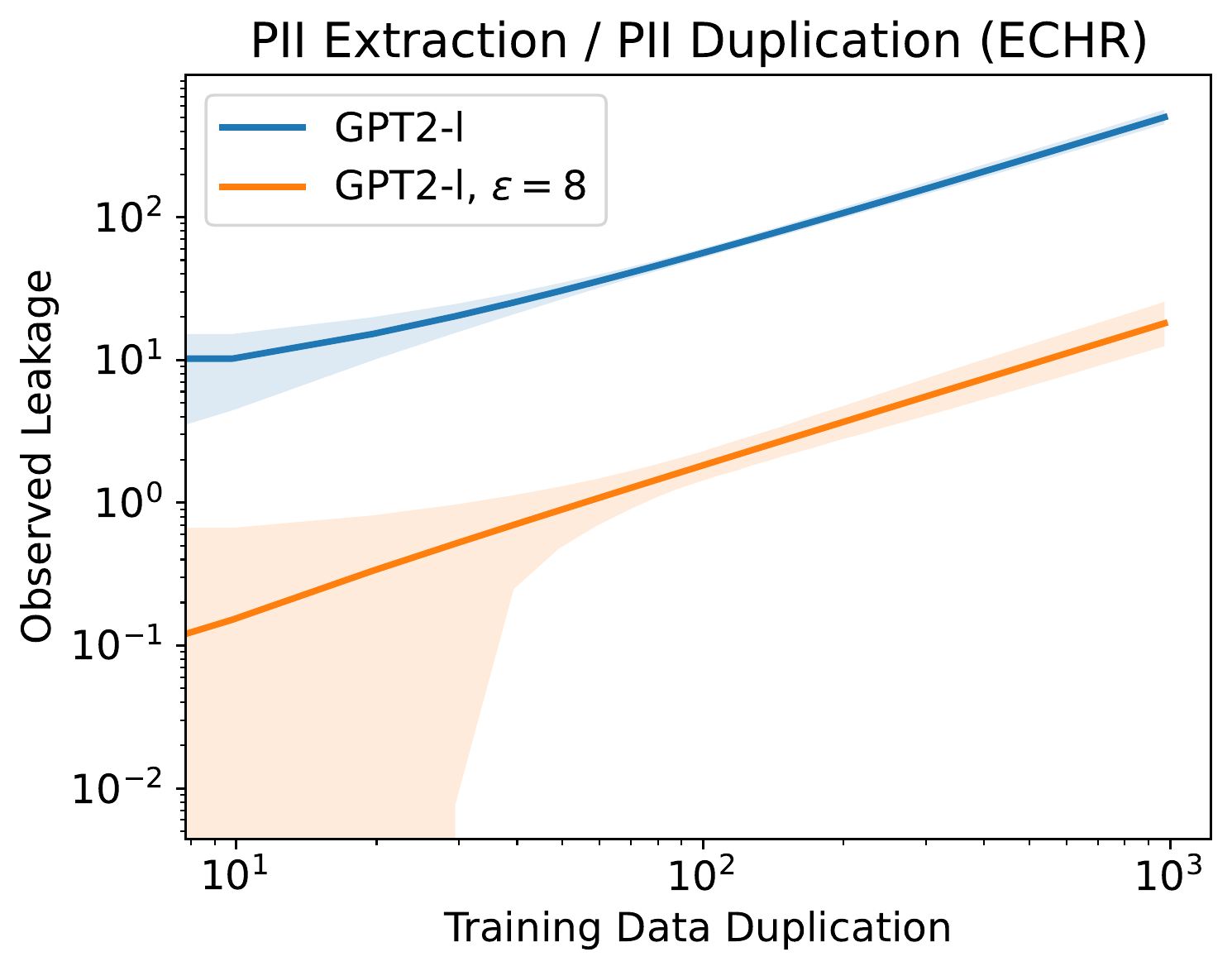} }}%
    \subfloat[\centering \label{subfig:ex_b}]{{\includegraphics[width=.24\linewidth]{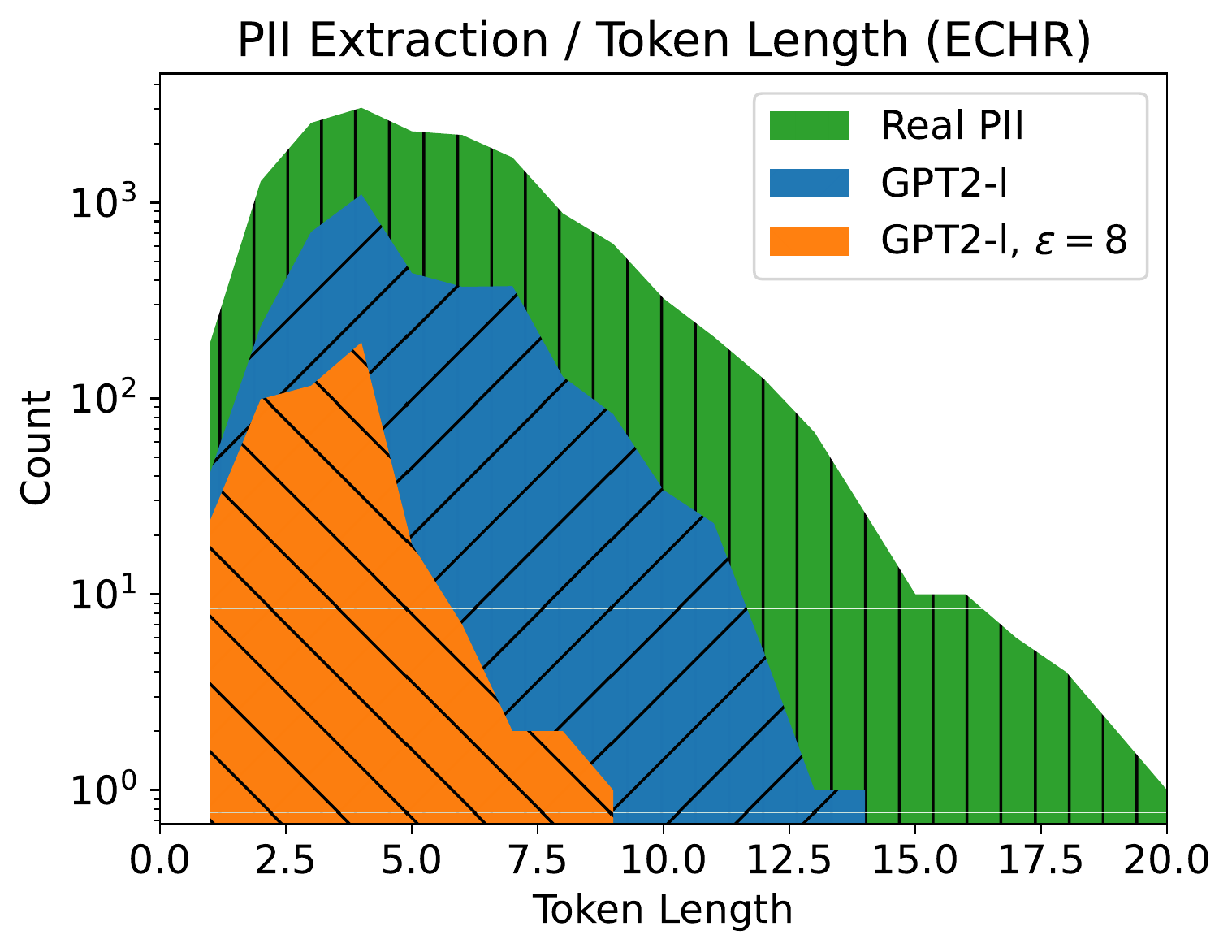} }}%
    \subfloat[\centering \label{subfig:ex_c}]{{\includegraphics[width=.24\linewidth]{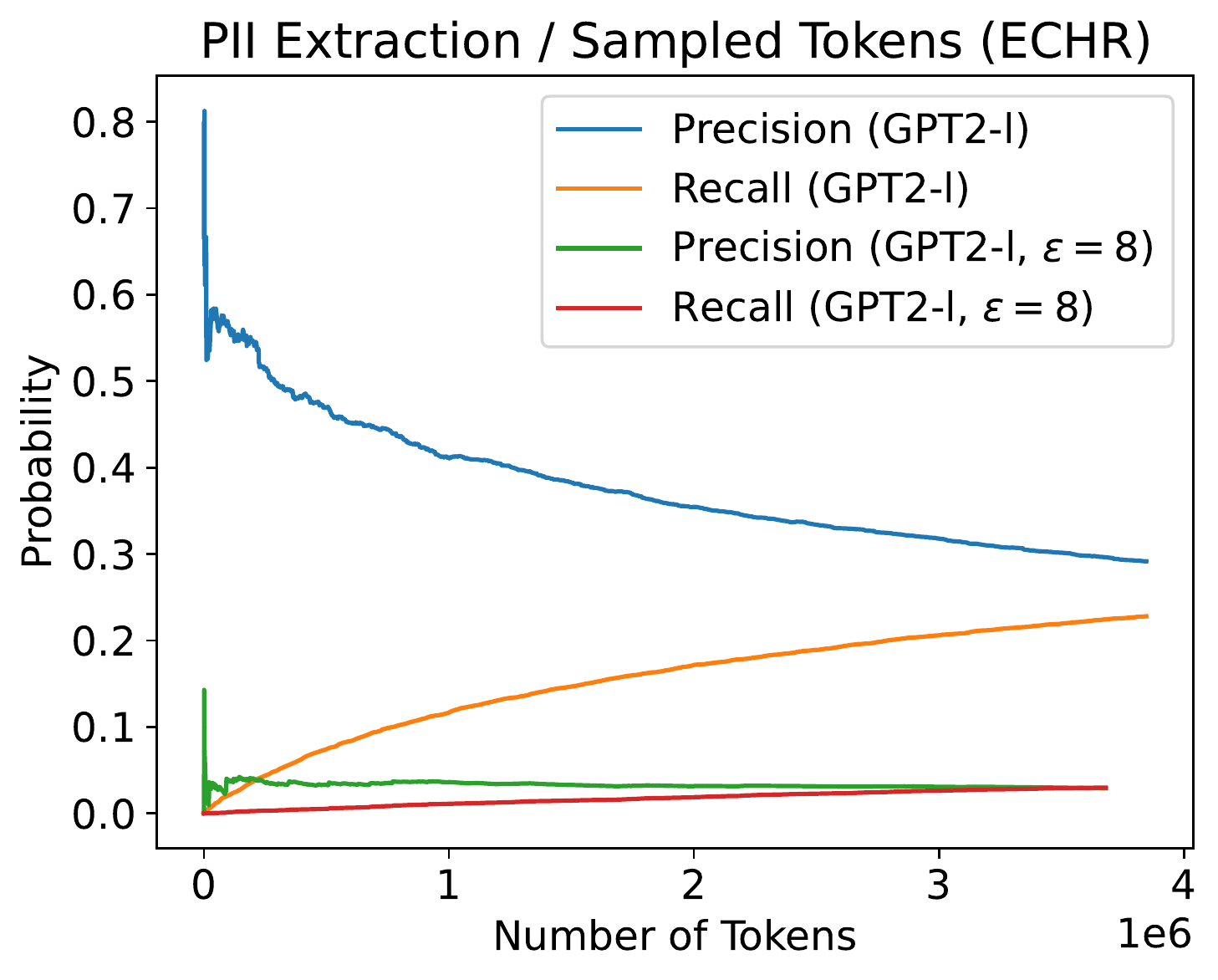} }}
    \subfloat[\centering \label{subfig:ex_d}]{{\includegraphics[width=.25\linewidth]
    {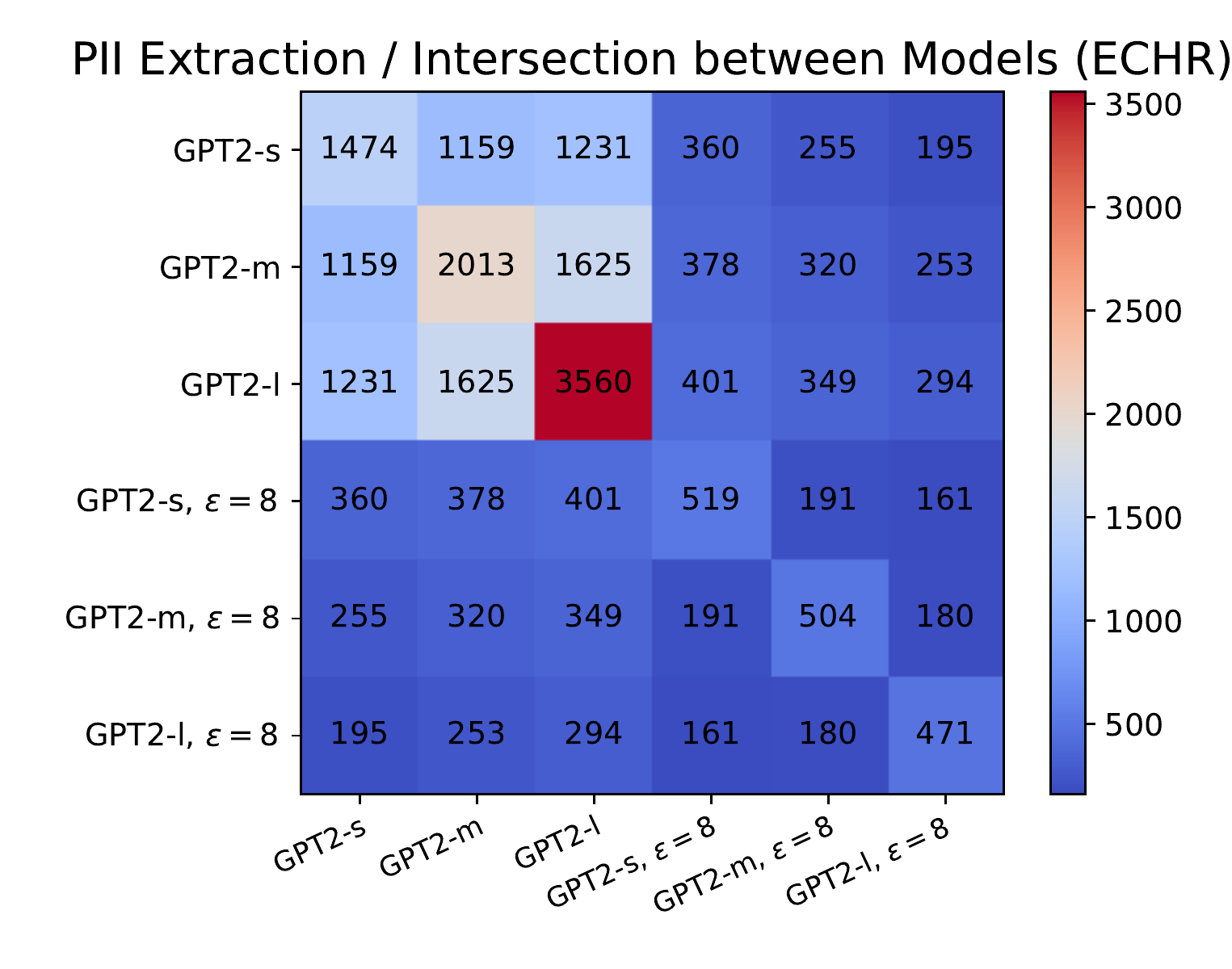} }} 
    \caption{A study of the PII that can be extracted with our attack after sampling about 4m tokens. 
    \cref{subfig:ex_a} shows that PII duplication strongly predicts leakage. 
    \cref{subfig:ex_b} shows that DP protects PII consisting of many tokens against extraction. 
    The ``Real PII" represent all raw PII from the same class in the training data. 
    \cref{subfig:ex_c} shows precision and recall as a function of the amount of tokens sampled for DP and non-DP models. 
    \Cref{tab:pii_extraction} details these results for different model sizes.
    \cref{subfig:ex_d} shows the intersection of extracted PII between models. The diagonal shows the number of extracted PII for each model.}
    \label{fig:results_extraction} 
\end{figure*}

\subsection{Metrics}
\label{sec:metrics}

We report the following metrics for measuring (i) model utility, (ii) vulnerability to membership inference (MI), and (iii) PII leakage.  
\begin{itemize}
    \item \textbf{Utility}: We compute the perplexity over an unseen test set, similar to related works~\cite{carlini2022quantifying, brown2022does, inan2021privacy}. 
    \item \textbf{Membership Inference}: We report the ROC AUC to measure sentence-level membership inference. 
    \item \textbf{PII Extractability}: We report Precision and Recall on the set of extractable PII.
    Recall measures (i) how much PII is at risk of extraction and Precision measures (ii) an attacker's confidence that a generated PII appears in the training dataset.
    \item \textbf{PII Reconstruction and Inference}: 
    We report the top-1 accuracy of correctly predicting a PII for a context. 
\end{itemize}
We refer to \Cref{sec:appendix_metrics} for formal definitions of these metrics. 

\subsection{PII Extraction}
\label{sec:pii_extraction}

We first extract PII from LMs trained with and without DP using \Cref{alg:observed_extraction}. 
Then, we show that the estimated leakage (from \Cref{alg:estimated_extractability}) matches the observed leakage which allows an attacker to point-wise verify the extractability of a PII without sampling the model exhaustively (making our attack more efficient). We analyse different factors such as duplication rate, token length, and sample size for their effect on PII leakage.

\autoref{tab:pii_extraction} shows the measured precision and recall with an ablation over the LM's size.
We sample on the order of 4m tokens from each target LM, by issuing 15k queries requesting the LM to generate sequences with a length of 256 tokens from an empty prompt using top-$k$ sampling with $k=40$. 
We account for baseline leakage by excluding all PII occurring in a random sample of 13m tokens in 50k queries from a public model (see \Cref{sec:baseline_leakage}).

\begin{table}[t]
\caption{Results for the observed PII extraction on ECHR (top rows), Enron (middle rows), and Yelp-Health (bottom rows)  after sampling around 4m tokens across 15k queries.}
\begin{tabular}{lllllll}
\hline
          & \multicolumn{2}{c}{\textbf{GPT2-Small}} & \multicolumn{2}{c}{\textbf{GPT2-Medium}} & \multicolumn{2}{c}{\textbf{GPT2-Large}} \\ \hline
          & No DP       & $\varepsilon=8$     & No DP        & $\varepsilon=8$     & No DP       & $\varepsilon=8$     \\ \hline \rowcolor{gray!13} 
          & \multicolumn{6}{c}{\textbf{ECHR}} \\ 
Prec & 24.91\%     & 2.90\%           & 28.05\%      & 3.02\%            & 29.56\%     & 2.92\%           \\
Recall    & 9.44\%      & 2.98\%           & 12.97\%      & 3.21\%           & 22.96\%     & 2.98\%           \\  \rowcolor{gray!13} 
    & \multicolumn{6}{c}{\textbf{Enron}} \\ 
Prec &  33.86 \% &  9.37\%    & 27.06\%    & 12.05\%   & 35.36\%    & 11.57\%           \\ 
Recall    &  6.26\%    & 2.29\%        & 6.56\%    & 2.07\%       & 7.23\%   & 2.31\%          \\
\rowcolor{gray!13} 
    & \multicolumn{6}{c}{\textbf{Yelp-Health}} \\
Prec & 13.86\%  &  8.31\%          & 14.87\%      & 6.32\%  & 14.28\%      & 7.67\%          \\
Recall    & 11.31\%      &  5.02\%   & 11.23\%   &  5.22\%  & 13.63\%     & 6.51\%           \\ \hline
\end{tabular}
\label{tab:pii_extraction}
\end{table}

\textbf{Model Size.} 
We observe that GPT-2-Large recalls 23\% of PII in the ECHR dataset with a precision of 30\%.
We conclude that in practice an attacker can be confident that a generated PII is contained in the training dataset. 
The precision and recall decrease with the model's size.
The smallest model (GPT-2-Small) has a significantly lower recall (only about 9\%) at a similar precision as the large models (25\%).  
In models trained with DP on ECHR, the precision and recall are similar for all model architectures, where we can extract about 3\% of PII with a precision of 3\%.

\textbf{Duplication.} 
\Cref{subfig:ex_a} shows that duplication of PII has a strong impact on their extractability. 
We group all PII with equal duplication count in the training dataset and compute the frequency with which they are generated by the model. 
On all datasets, we observe a linear relationship between the number of occurrences of a piece of PII and the frequency with which they are leaked, \ie, PII occurring twice as often is expected to also leak twice as often. 
Our finding contrasts with the \emph{superlinear} effect in sequence-level memorization observed by \citet{kandpal2022deduplicating}.
We note that \citet{kandpal2022deduplicating} study models trained from scratch and arbitrary sequences, whereas our study focuses on fine-tuned models and PII. Further examination is necessary to fully comprehend the relationship between both findings.
In DP models, we observe that the extractability of PII is consistently about an order of magnitude lower than in undefended models. 

\textbf{Token Length.} 
We compare leaked PII by token length to evaluate whether PII sequences containing more tokens are less prone to extraction. 
In Figure~\ref{subfig:ex_b}, we group all leaked PII by their token length and compute a mean count from the generated dataset. 
We observe that undefended models leak PII sequences containing many tokens whereas long sequences are not leaked in DP models. 
In the range between 3-6 tokens, we observe that DP models leak about an order of magnitude fewer pieces of PII than undefended models. 

\textbf{Sample Size.} 
\Cref{subfig:ex_c} shows precision and recall as a function of the number of sampled tokens on ECHR. 
The recall increases to 23\%, whereas the precision consistently decreases from 50\% at 500k tokens to 30\% at 4m tokens. 
This indicates that PII with a high probability of generation by the LM are likely pieces of real PII from the dataset and thus vulnerable to extraction. 
An attacker who samples larger sets from the model can generate more PII, but at a lower precision which makes the attack less impactful. 

\begin{figure*}%
    \centering
    \subfloat[\centering \label{subfig:observed_leakage}]{{\includegraphics[width=.32\linewidth]
    {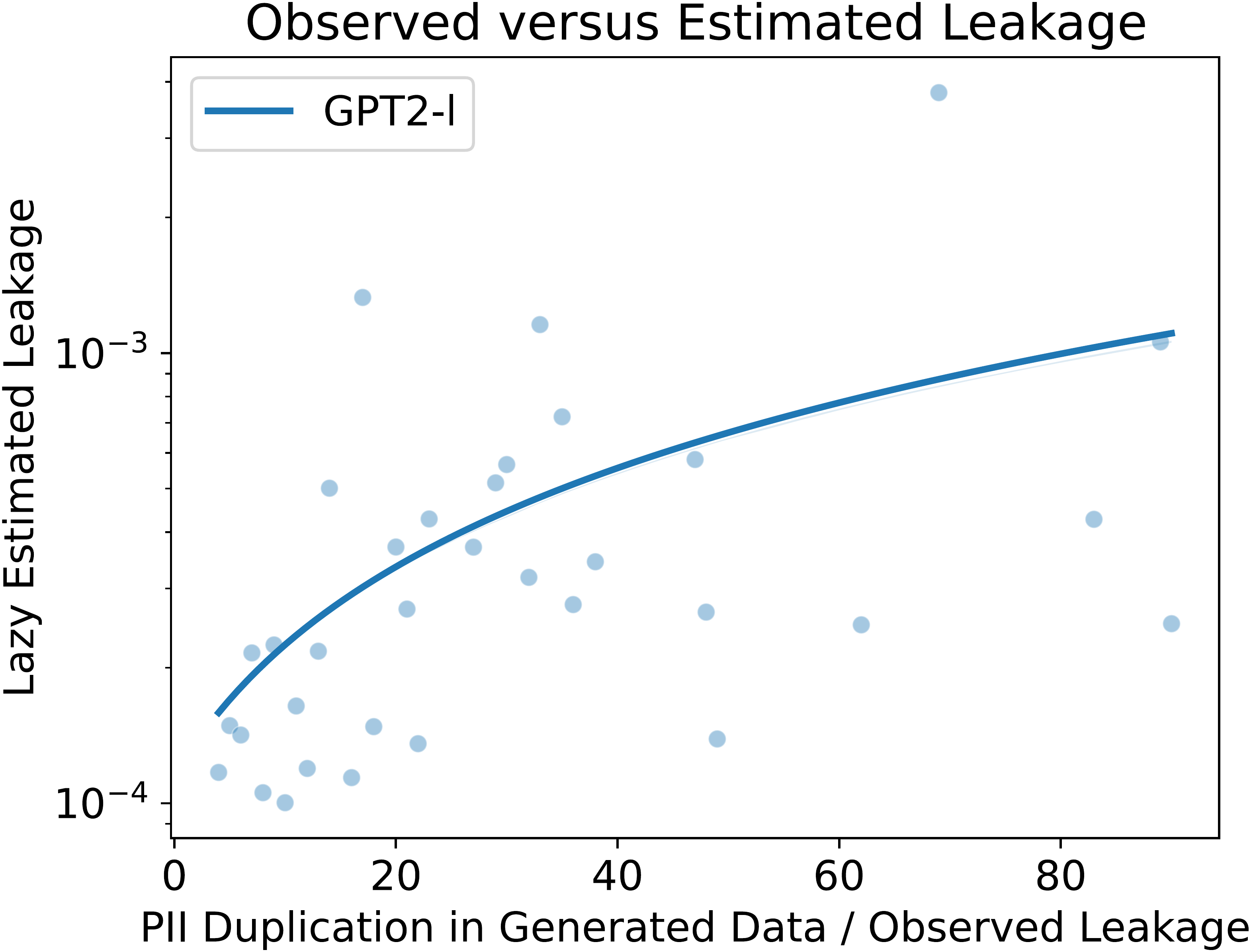} }} 
    \subfloat[\centering \label{subfig:ex2_a}]{{\raisebox{-2pt}{\includegraphics[width=.313\linewidth]{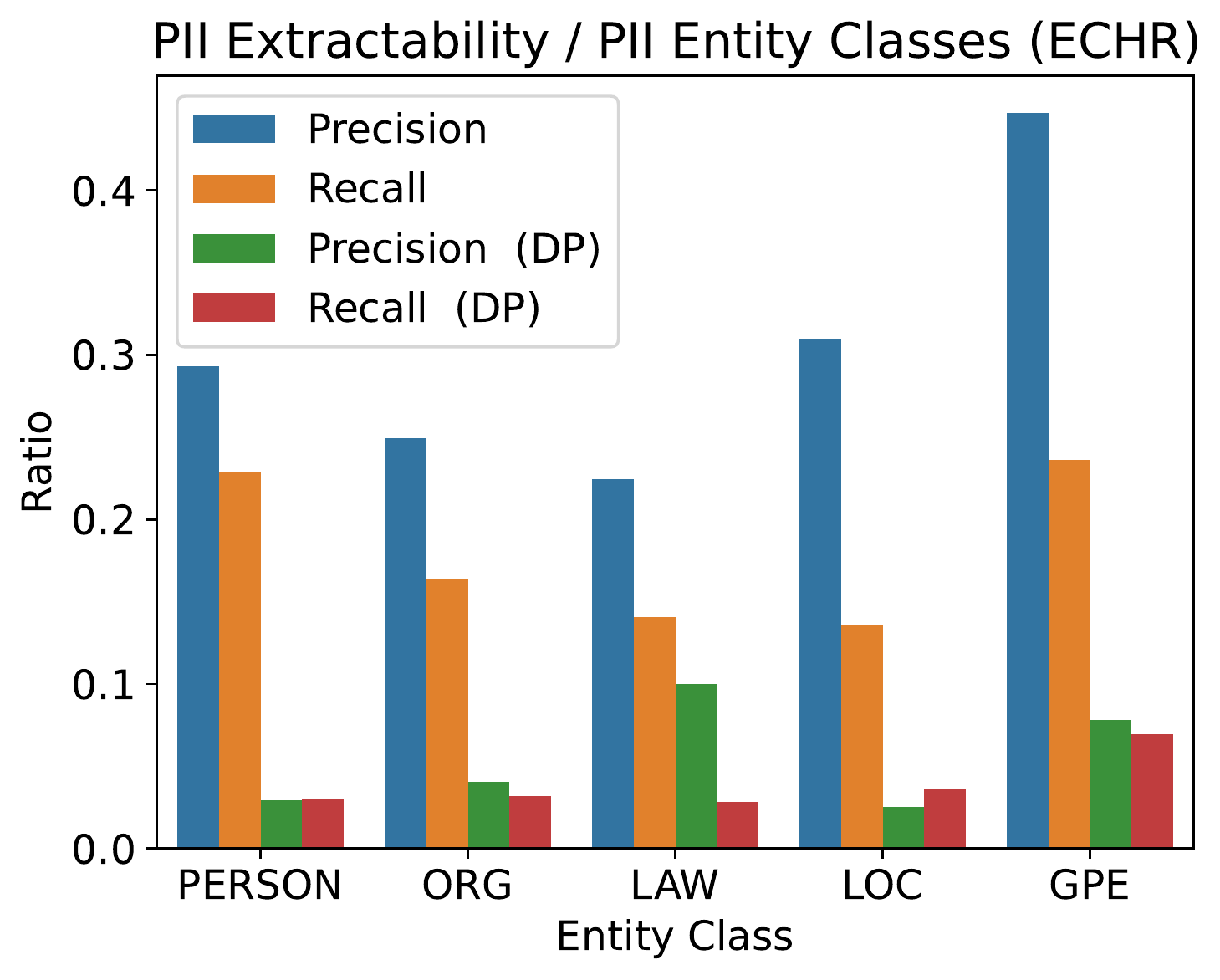}} }}%
    \subfloat[\centering \label{subfig:inference_context}]{{\raisebox{-2pt}{\includegraphics[width=.32\linewidth]{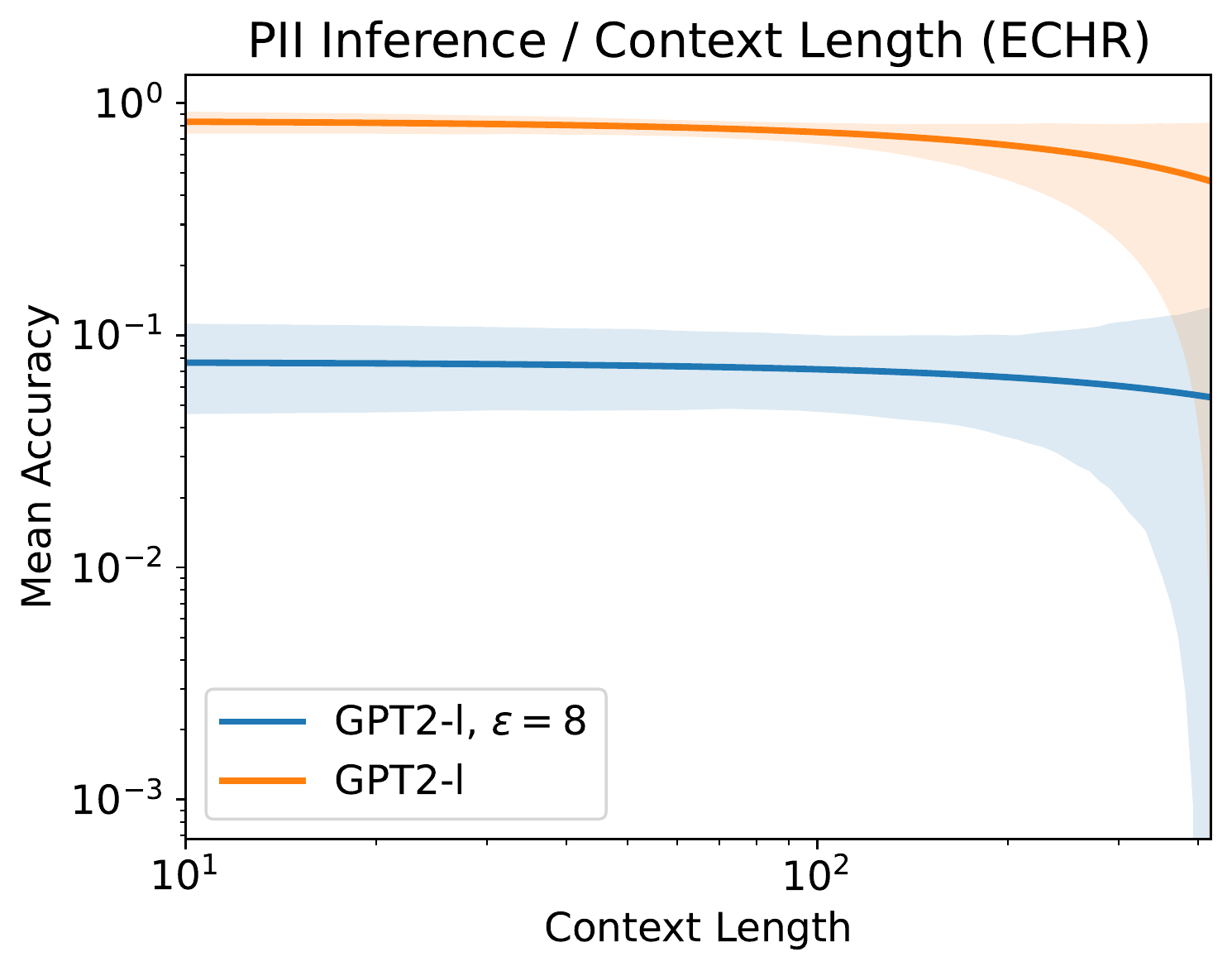}} }}
    \caption{
      \cref{subfig:observed_leakage} shows the correlation between the observed and estimated leakage. \cref{subfig:ex2_a} shows the precision and recall for other entity classes on the ECHR dataset. \cref{subfig:inference_context} shows the mean inference accuracy relative to the context length, which is the length of the combined prefix and suffix for a masked query. 
    }
    \label{fig:results_metrics} 
\end{figure*}

\textbf{Similarities between Generated PII.} 
Figure~\ref{subfig:ex_d} shows the intersection between sets of extracted PII across all LMs in a heatmap.
We observe that (i) a subset of the most duplicated PII occurs in almost all and (ii) there is little similarity between which PII were leaked between models. 
In undefended models, we observe that PII which occur a single time can leak, which we never observe for DP models. 

\textbf{Estimated Extractability.} \cref{subfig:observed_leakage} shows that lazily estimated extractability correlates with observed leakage (\ie, the number of times a model generates the PII).
This allows computing point-wise estimates for a target PII without searching through massive amounts of generated text. 
Our metric is however not perfect, as some false negative outliers are incorrectly predicted as non-extractable despite appearing in the generated dataset. 

\begin{table*}[t]
    \centering
    \begin{tabular}{@{}lllllllll@{}}
\toprule
                                                                   & \multicolumn{2}{c}{\textbf{GPT2-Small}}                       & \multicolumn{2}{c}{\textbf{GPT2-Medium}}                      & \multicolumn{2}{c}{\textbf{GPT2-Large}}                       & \multicolumn{2}{c}{\textbf{GPT2-XL}}                                  \\ \midrule
\multicolumn{1}{l}{}                                              & \multicolumn{1}{c}{No DP} & \multicolumn{1}{c}{$\varepsilon=8$} & \multicolumn{1}{c}{No DP} & \multicolumn{1}{c}{$\varepsilon=8$} & \multicolumn{1}{c}{No DP} & \multicolumn{1}{c}{$\varepsilon=8$} & \multicolumn{1}{c}{No DP} & \multicolumn{1}{c}{$\varepsilon=8$} \\ \midrule
\multicolumn{1}{l}{ECHR(TAB)}                   & 0.78\%                    & \multicolumn{1}{l}{0.24\%}       & 1.21\%                    & \multicolumn{1}{l}{0.32\%}       & 5.81\%                    & \multicolumn{1}{l}{0.48\%}       & 4.30\%                    & 0.39\%                           \\
\rowcolor{gray!13} \multicolumn{1}{l}{ECHR (Ours, $|\mathcal{C}|=64$)}         & \textbf{2.25\%}           & \multicolumn{1}{l}{0.44\%}       & \textbf{3.36\%}           & \multicolumn{1}{l}{0.87\%}       & \textbf{18.27\%}          & \multicolumn{1}{l}{0.55\%}       & \textbf{13.11\%}          & 0.41\%                  \\
\multicolumn{1}{l}{Enron (TAB)}                 & 0.59\%                    & \multicolumn{1}{l}{0.04\%}       & 0.67\%                    & \multicolumn{1}{l}{0.04\%}             & 1.75\%                    & \multicolumn{1}{l}{0.04\%}              & 2.19\%                    &  0.19\%                                \\
\rowcolor{gray!13} \multicolumn{1}{l}{Enron (Ours, $|\mathcal{C}|=64$)}       & \textbf{6.29\%}           & \multicolumn{1}{l}{0.49\%}       & \textbf{7.26\%}           & \multicolumn{1}{l}{0.52\%}             & \textbf{12.68\%}          & \multicolumn{1}{l}{0.55\%}             & \textbf{15.25\%}          &  0.53\%                                     \\
\multicolumn{1}{l}{Yelp-Health (TAB)}           & 0.33\%                    & \multicolumn{1}{l}{0.24\%}       & 0.37\%                    & \multicolumn{1}{l}{0.14\%}       & 0.65\%                    & \multicolumn{1}{l}{0.12\%}       & 1.99\%                    & 0.12\%                           \\
\rowcolor{gray!13} \multicolumn{1}{l}{Yelp-Health (Ours, $|\mathcal{C}|=64$)} & \textbf{0.42\%}           & \multicolumn{1}{l}{0.32\%}                            & \textbf{1.31\%}           & \multicolumn{1}{l}{0.32\%}                            & \textbf{1.69\%}           & \multicolumn{1}{l}{0.35\%}                            & \textbf{6.40\%}           & 0.36\%                  \\ \bottomrule
\end{tabular}
    \caption{Results of PII reconstruction attacks on the entity class ``person". 
    Bold numbers represent the best attack per dataset and LM. 
    We compare our results with the TAB attack~\cite{inan2021privacy} on three datasets.
    \label{tab:reconstruction_results}}
\end{table*}

\textbf{Extraction of other PII Classes.} We measure the PII extractability for other classes of PII such as e-mails and phone numbers. 
The attacker can extract about 14.1\% of law case numbers and 16.3\% of mentioned organization names from an undefended model (shown in \Cref{subfig:ex2_a}). 
In the DP model, we observe that an attacker can only extract 2.8\% of law cases and 4.1\% of organizations. 
For the Enron dataset, which contains long phone numbers, we never observe a single leaked real phone number in the DP model. 
However, we observe leakage of e-mail addresses (consisting of equally many tokens), that are typically correlated with a person's name. 

\subsection{PII Reconstruction}
We compare our PII reconstruction attack from \Cref{alg:estimated_reconstruction} with the TAB attack~\cite{inan2021privacy}. 
\Cref{tab:reconstruction_results} shows the results on ECHR, Enron, and Yelp-Health for the entity class `person'.
We sample $64$ candidates and decode from the model using top-$k$ sampling with $k=40$.
We observe that our reconstruction attack significantly outperforms the TAB attack on undefended models enabling the reconstruction of up to $10 \times$ more PII (in the GPT-2-Medium case on Enron).

\textbf{Model Size.} 
On ECHR and GPT-2-Large, TAB correctly reconstructs at least $5.81\%$ of PII whereas our attack achieves $18.27\%$. 
This observation demonstrates that information in a sample's suffix provides a strong signal to reconstruct PII. 
On ECHR, our attack improves the baseline by at least $2.5 \times$, on Enron we observe an improvement of at least $7.5\times$ and on Yelp-Health our attack is at least about $3 \times$ more successful (except for GPT-2-Small where our attack improves only from 0.33\% to 0.42\%). 
The risk of reconstruction is much smaller in DP models ($\leq 1\%$) where our attack still improves the baseline in all cases, but we believe the leakage is too small for a practical attack.  
We observe that across all datasets, larger models are more vulnerable to PII reconstruction.

\textbf{Context Size.} 
On Enron, the advantage of our attack compared to TAB becomes more evident.
E-mails in the Enron dataset typically mention the receiver of the e-mail at the beginning prior to any PII. 
For this reason, the TAB attack has only a small prefix to predict PII and cannot leverage the information contained in the e-mail's body. 
We observe that when the PII is in the set of candidates, it is predicted correctly about 70\% of the time.
However, our reconstruction attack often does not sample the correct candidate which effectively limits our attack's success rate.
We believe a method that samples candidates by incorporating information from the sample's suffix could improve our attack even further. 

\begin{table}[ht]
    \centering
    \caption{Results of our PII inference attack on fine-tuned versions of GPT-2-Large. 
    The values represent the attack's accuracy at inferring the correct PII out of $|\mathcal{C}|$ candidates.}
    \begin{tabular}{@{}lllllll@{}}
\toprule
      & \multicolumn{2}{c}{\textbf{ECHR}} & \multicolumn{2}{c}{\textbf{Enron}} & \multicolumn{2}{c}{\textbf{Yelp-Health}} \\ \midrule
      & No DP        & $\varepsilon=8$       & No DP         & $\varepsilon=8$       & No DP           & $\varepsilon=8$           \\\midrule

$|\mathcal{C}|=100$ &  70.11\%         &   8.32\%        & 50.50\%       & 3.78\%             &      28.31\%           &     4.29\%                   \\\rowcolor{gray!13} 
$|\mathcal{C}|=500$ &     51.03\%         &    3.71\%           &  34.14\%             &   1.92\%                 &      15.55\%           &      1.86\%                  \\ \bottomrule
\end{tabular}
    \label{tab:inference_results}
\end{table}

\subsection{PII Inference}
In PII inference, our attacker has access to (i) the anonymized training dataset and (ii) a list of candidate PII that also appear in the training dataset. 
For PII inference, we evaluate our attacks against the GPT-2-Large model on all three surveyed datasets with and without DP.
Table~\ref{tab:inference_results} summarizes the results from our attack.

\textbf{Results.} We observe that in the undefended setting, an attacker can infer PII with an accuracy of 70\% out of 100 candidates on ECHR, 50\% on Enron, and 28\% on Yelp-Health. 
We observe higher leakage on ECHR and Enron, which is likely because they have more structure than Yelp reviews. 
Pieces of PII are mentioned repeatedly in similarly structured sentences, which causes higher PII leakage. 
The undefended setting enables practical attacks where the attacker can be confident about results. 
In the DP setting, an attacker can achieve an accuracy of about 8\% given 100 candidates and about 4\% in 500 candidates on ECHR. 
Although leakage in DP models is small, we believe our experiments demonstrate that DP does not fully protect against PII leakage in a practical setting against an informed attacker. 
\cref{subfig:inference_context} shows that inferring PII in very large contexts slightly \emph{worsens} the accuracy. 
This is likely because the expected memorization per token is lower in samples containing many tokens. 

\subsection{Membership Inference and PII Leakage}
\label{sec:connection}
\begin{figure*}%
    \centering
    \subfloat[\centering \label{subfig:roc_auc}]{{\includegraphics[width=.34\linewidth]{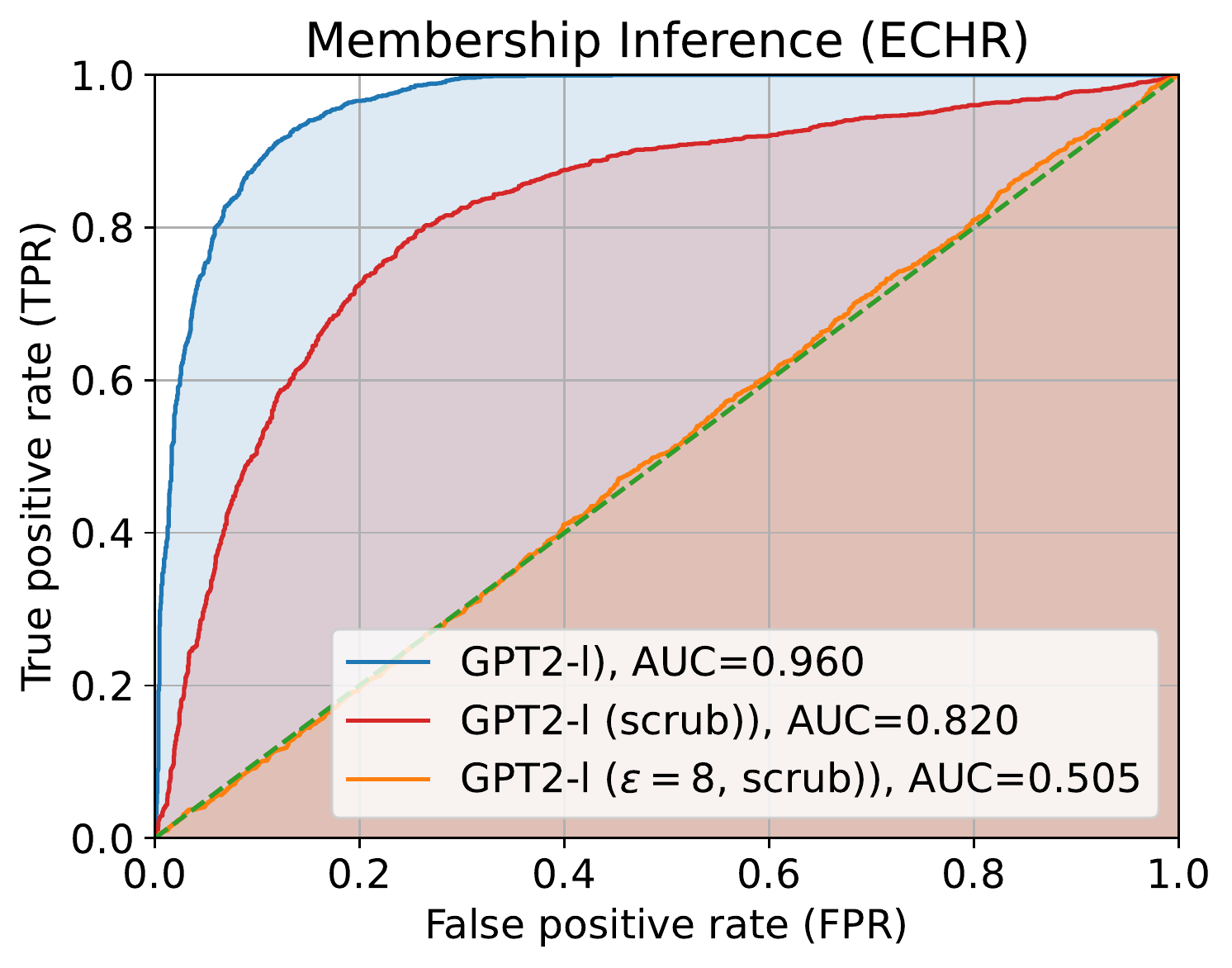} }} 
    \subfloat[\centering \label{subfig:mem_vs_recon}]{{\includegraphics[width=.27\linewidth]{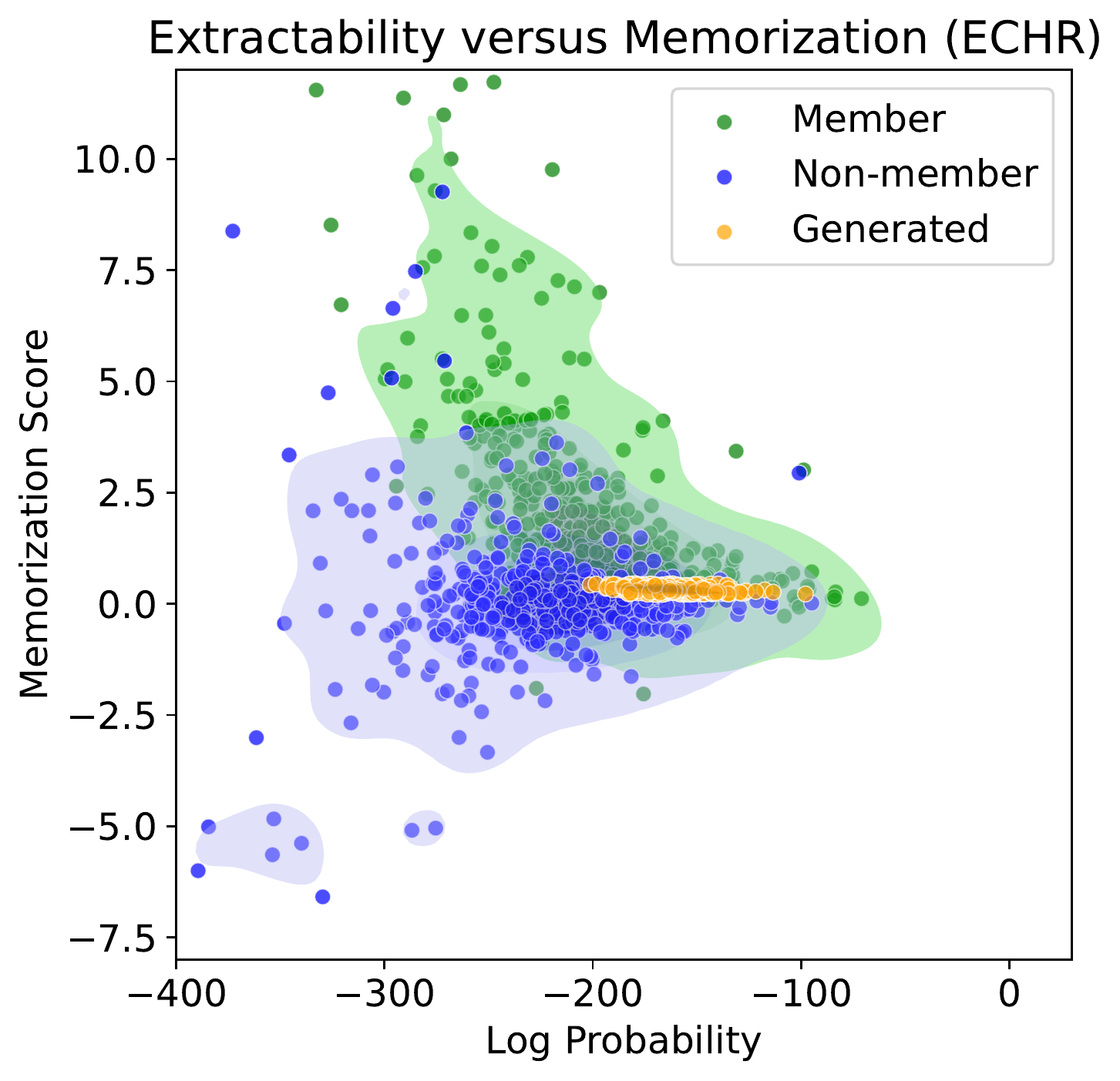} }}%
    \subfloat[\centering \label{subfig:memorization}]{{\includegraphics[width=.34\linewidth]{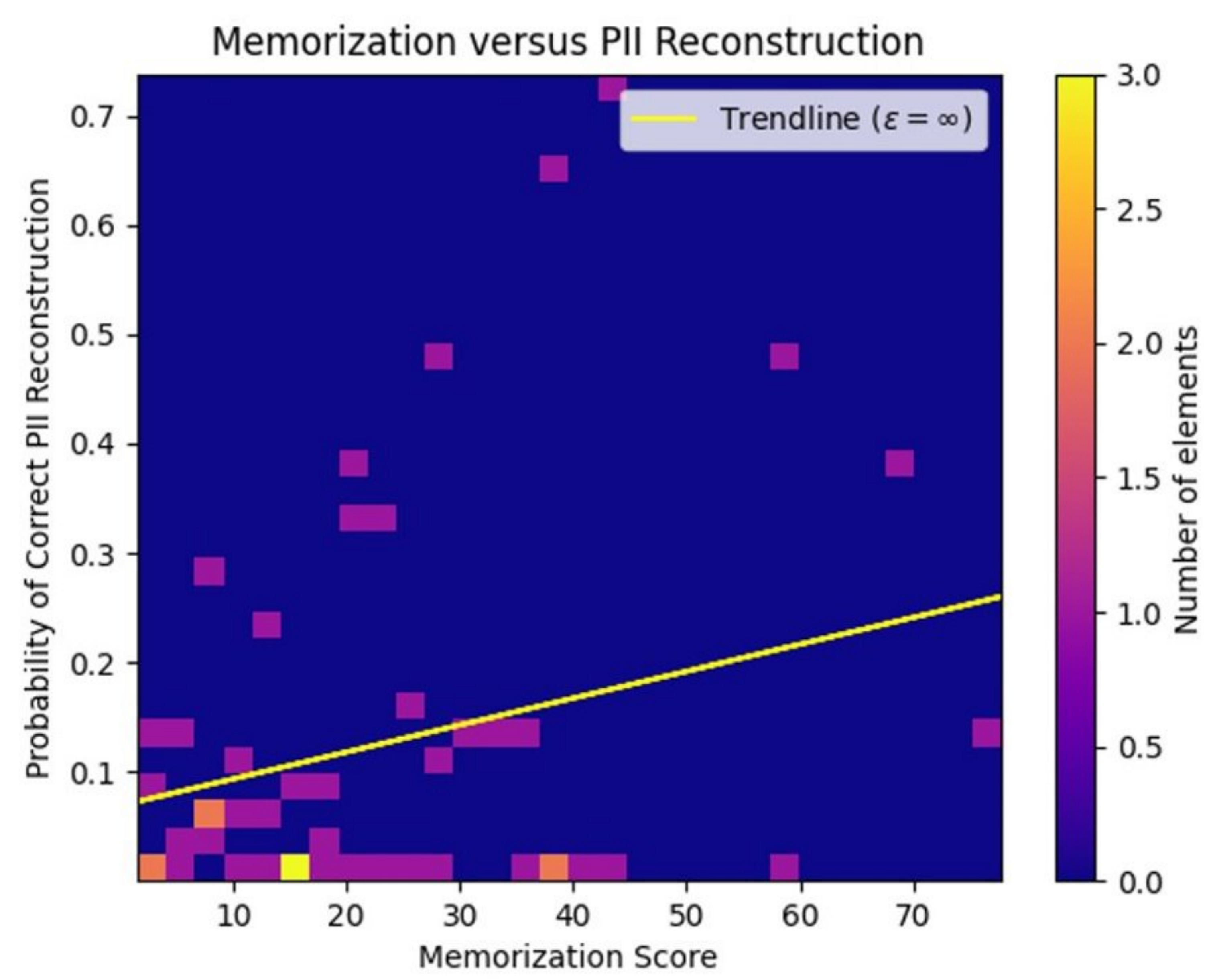} }}%
    \caption{
    Connecting sentence-level membership inference with PII reconstruction in GPT-2-Large. \ref{subfig:roc_auc} shows the ROC curve against our fine-tuned model using a shadow model attack on ECHR.
    \ref{subfig:mem_vs_recon} shows that the memorization score of generated sequences is nearly zero and \ref{subfig:memorization} shows that the memorization score correlates with the probability of correct PII reconstruction. }
    \label{fig:results_mia} 
\end{figure*}

We employ a shadow model membership inference attack~\cite{shokri2017membership} to empirically evaluate the relationship between sentence-level membership inference and PII leakage.
In this attack, the adversary trains shadow models on datasets sampled from the same data distribution as the target model.
The adversary then calculates the difference between the perplexity (\ie $\textsc{PPL}$) of the target sentence \wrt the target and shadow models, and uses this as a score to decide if the sentence was a training member or not.

\cref{subfig:roc_auc} shows the ROC curve of this MI attack against an undefended model, a model trained after scrubbing, and a model trained with differential privacy after scrubbing. 
Expectedly, we observe that scrubbing PII mitigates membership inference, but is nowhere as effective as DP.
The attack achieves an AUC score of 0.96, 0.82, and 0.505 against undefended, scrubbed, and DP \& scrubbed models respectively.

\textbf{Connection between MI and PII Leakage.}
\Cref{alg:inference_game_indistinguishability} shows the sentence-level MI game of \citet{yeom2020overfitting} alongside an indistinguishability variant of the PII Inference game in \Cref{alg:inference_game}. This corresponds to the case $m=1$ when the adversary would be given a pair of candidates for the target PII, with the only exception that in \Cref{alg:inference_game_indistinguishability} the adversary is given just one of the candidates, depending on a Bernoulli random variable $b$.
The difference between one or two candidates is inessential and analogous variants of sentence-level MI have been considered before~\cite{humphries2020differentially}.
The essential difference between the MI and PII Inference games is that in the former the adversary has to distinguish between two sentences sampled from $\mathcal{D}$, while in the PII Inference game, it has to distinguish between two sentences differing only on a piece of PII.
This means that to leverage a MI attack for PII reconstruction or inference, the MI attack has to be strong enough to distinguish between member and non-member sentences differing in a few tokens. In contrast, a PII Inference attack can be directly turned into a MI attack against sentences containing a piece of PII.

PII Reconstruction is similarly analogous to attribute inference (where the target \emph{attribute} is the missing piece of PII).
PII Reconstruction can be linked to MI the same way as attribute inference was linked to MI by \citet{yeom2020overfitting}. 
Our empirical results show that they are indeed correlated. 
\begin{algorithm}[t]
\caption{Sentence-level MI (lines enclosed in solid box) vs. PII Inference (lines enclosed in dashed box).}
\begin{algorithmic}[1]
\game{Ind-Inference}{$\mathcal{T}, \mathcal{D}, n, \mathcal{A}$} 
    \State $b \sim \{0,1\}$
    \State $D \sim \mathcal{D}^n$
    \State $\theta \gets \mathcal{T}(D)$
    \BeginBox
    \State $S_0 \sim D$
    \State $S_1 \sim \mathcal{D}$
    \State $\tilde{b} \gets \mathcal{A}(\mathcal{T}, \mathcal{D}, n, \mathcal{O}_\theta(\cdot), S_b)$
    \EndBox
    \BeginBox[draw=black,dashed]
    \State $S \sim \{S \in D| \textsc{Extract}(S) \neq \emptyset\}$
    \State $C_0 \sim \textsc{Extract}(S)$
    \State $C_1 \sim \mathcal{E}$
    \State $\tilde{b} \gets \mathcal{A}(\mathcal{T}, \mathcal{D}, n, \mathcal{O}_\theta(\cdot), \textsc{Scrub}(\textsc{Split}(S,C_0)), C_b)$
    \EndBox
\EndGame
\end{algorithmic}
\label{alg:inference_game_indistinguishability}
\end{algorithm}
For instance, \cref{subfig:mem_vs_recon,subfig:memorization} respectively contrast MI with PII extractability and reconstruction attacks. 
\cref{subfig:mem_vs_recon} shows the likelihood of the model producing a member on the $x$-axis versus the memorization score on the $y$-axis.
We observe that samples generated from empty prompts are not memorized, meaning that they likely contain few signals that are useful for MI. 
\cref{subfig:memorization} shows the memorization score relative to our reconstruction attack. 
We observe a positive correlation between a sentence's memorization score and the success rate of our PII reconstruction attack. 
This means that sentences that are vulnerable to the MI attack are usually vulnerable to the PII reconstruction attack, and vice-versa.

\subsection{Summary of Results}

\Cref{tab:summary_results} summarizes the privacy-utility trade-off for GPT-2-Large model when fine-tuned on ECHR dataset. We enumerate our key results below:

\begin{itemize}
    \item Undefended models are highly vulnerable to all privacy attacks including membership inference and PII extraction, reconstruction, and inference. For PII risks, larger model sizes and higher duplication counts increase the risk of leakage.   
    \item Our results show that the threat of reconstructing PII has been underestimated and we demonstrate up to an order of magnitude higher leakage than prior work.
    \item Empirically, we observe that differential privacy significantly bounds leakage from PII reconstruction relative to undefended models. 
    \item DP does not completely eliminate leakage from PII inference and PII extraction. 
    We demonstrate that an attacker can infer PII with up to 10\% accuracy (given 100 candidates) in a practical setting.  
    \item We find that DP and (aggressive) PII scrubbing limit the LM's utility, motivating the search for defenses with better empirical privacy/utility trade-offs.
\end{itemize}

\begin{table}[t]
    \centering
    \begin{tabular}{@{}l@{}cccc@{}}
\toprule
                  & \textbf{Undefended} & \textbf{DP} & \textbf{Scrub} & \textbf{DP + Scrub} \\ \midrule
Test Perplexity        & 14 / 9              & 14           &  16            & 16                   \\\rowcolor{gray!13}
Extract Precision & 30\%           & 3\%         & 0\%            & 0\%                 \\
Extract Recall    & 23\%           & 3\%         & 0\%            & 0\%                 \\\rowcolor{gray!13}
Reconstruction  Acc.  & 18\%           & 1\%         & 0\%            & 0\%                 \\
Inference Acc. ($|\mathcal{C}|=100$) & 70\%           & 8\%         & 1\%            & 1\%                 \\\rowcolor{gray!13}
MI AUC           & 0.96           & 0.5         & 0.82           & 0.5                 \\ \bottomrule
\end{tabular}
    \caption{Our results on ECHR for GPT-2-Large summarize the privacy-utility trade-off. 
    We show the undefended model's perplexity with/without masking generated PII. 
    The undefended model has the lowest perplexity but the highest leakage. DP with $\epsilon=8$ mitigates MI and (partially) PII leakage. 
    Scrubbing only prevents PII leakage.
    DP with scrubbing mitigates all the privacy attacks but suffers from utility degradation.}
    \label{tab:summary_results}
\end{table}

\section{Discussion and Limitations}
\label{sec:discussion}
Below, we discuss extensions and limitations of our methodology, and identify further research motivated by our findings.
We first discuss the applicability of our methodology to sensitive information other than PII, and potential extensions to our attacks exploiting semantic similarity and associations in the training dataset.
We then describe how masked language models fare compare to autoregressive models and identify further research motivated by our findings: how to best combine DP training and scrubbing, optimizing attacks for other leakage metrics, and the need for better benchmarks.

\textbf{General Applicability.} 
In this paper, we focus on defining metrics, game-based definitions, and tractable formulas for evaluating leakage of sensitive sequences of tokens categorized as PII. That said, we bring attention to the point that our methodology is generally applicable to any notion of sensitive input. As long as one has an effective method to correctly identify inputs deemed sensitive, our methodology can be adapted to measure the protection offered by existing ML pipelines in mitigating the leakage of {\em any} sensitive information. In practice, it is often hard to draw a clear boundary around what constitutes sensitive information, which is an important but orthogonal problem.

\textbf{Syntactic and Semantic Similarity.} 
We consider verbatim matches of PII tokens as leakage, however, our methods can be adapted to account for both syntactic and semantic similarity. For example, ``Mr. John Doe" and ``J. Doe" could be inferred to be the same person. Similarly, PII reconstruction and PII inference attacks can employ contexts with similar meaning to improve attack results.

\textbf{Advanced Attacks.} 
We consider leakage of PII sequences from the training dataset in isolation, irrespective of the context where it appears and other extracted PII. 
Extracted PII sequences can be further leveraged in advanced attacks that explore associations among them and reveal additional private information about the training dataset, thereby enabling linkability attacks.

\textbf{Utility-preserving Scrubbing.} 
Our empirical evaluation demonstrates that differential privacy is partially effective in mitigating leakage of PII. Based on this observation, existing scrubbing techniques can be adapted to take into consideration the partial protection offered by DP and heuristically scrub only PII that remains unprotected (\eg because it occurs many times).
Such a DP-informed scrubbing would allow for improving model utility while maintaing a privacy level equivalent to a naive combination of DP training and scrubbing.

\textbf{Comparison to Masked Language Models.}
Pior work has explored PII reconstruction in the clinical health setting~\cite{lehman2021does, vakili2021clinical} with masked language models (MLMs) based on the BERT architecture~\cite{devlin2018bert}.
MLMs are trained to reconstruct a word in a masked query, which is equivalent to the PII reconstruction task in \cref{eq:reconstruction}.
During training, BERT models optimize the masked word's probability conditioned on the prefix and suffix, in comparison to GPT-2 which is auto-regressive and can only be conditioned on the prefix.  
A service deploying an MLM enables trivial attacks to query the most-likely replacement for a single mask conditioned on \emph{the entire sentence}, unlike GPT-2 models. 
One of our contributions is to imitate this functionality in GPT-2 models to reconstruct PII. 

Attacks on GPT-2 and similar models employed for text generation are potentially a greater threat.
Autoregressive models typically expose a next-word prediction API, whereas BERT-like models are often used for downstream tasks such as text classification, with less revealing APIs.
We state that \cref{eq:reconstruction} is intractable, which is also true for existing MLMs since an attacker does not know the number of tokens in the PII and has to perform a general constrained search. 

\textbf{Need for Better Benchmarks.}
In conducting this research, we realized the shortage of good benchmark datasets for measuring PII leakage. A notable exception is the Text Anonymization Benchmark of \citet{pilan2022text} with human-annotated PII. However, we found this dataset to be too small for fine-tuning models with DP-SGD: the dataset contains a subset of $1,268$ out of the estimated $11\,500$ cases from the original ECHR dataset~\cite{Chalkidis:2019}. With such few records, we were unable to fine-tune models with both reasonable privacy (\ie, low $\varepsilon$) and utility (low perplexity).

Our work motivates the need for better benchmarks for evaluating attacks and mitigations against PII leakage in trained models, in addition to evaluating text anonymization techniques at a dataset level. In performing our evaluation, we rely on off-the-shelf text anonymization tools powered by NER models to tag PII in generated sentences and leverage them in computing our leakage metrics. As a consequence of this approach, our metrics can capture leakage that is dependent on the quality of the tools we use. Assuming that NER models and other techniques used in these tools are prone to error, our results provide a lower bound on the leakage.

\textbf{High-Precison/Low-Recall Attacks.} Our attacks evaluate PII leakage using average-case metrics and provide an overview of the threat of PII leakage in LMs.
We did not consider high-precision/low-recall attacks~\cite{carlini2022membership} and further research is needed to explore their potential impact and effectiveness.

\textbf{Limitations}. 
Due to the lack of extensive, annotated benchmark datasets for studying PII leakage in LMs, we employ the same NER model for both scrubbing and measuring PII leakage. \citet{pilan2022text} demonstrate that a fine-tuned NER model with ground-truth PII annotations achieves recall rates between 84-93\%, decreasing to 77\% without annotations. 
Since scrubbing cannot remove all PII and we show PII leakage empirically, we conclude that scrubbing cannot fully prevent PII leakage. 
Further research is required to expand our findings to other LMs and datasets.

\section{Related Work}
\label{subsec:related_work}
We discuss prior work on attacks inferring private data as well as defenses to mitigate the leakage.

\textbf{Extraction of Training Data.}
There is extensive work studying how large language models memorize training data and attacks inferring information under various threat models. Research has shown the feasibility of extracting different types of information including individual sentences~\cite{carlini2021extracting,inan2021privacy}, inserted canaries~\cite{carlini2019secret,parikh2022canary,zanella2020analyzing} as well as $n$-grams~\cite{mccoy2021much}.
Prior work studied the leakage of PII in masked language models~\cite{lee2022language,vakili2021clinical}, large language models~\cite{huang2022large,rocher2019estimating} and Smart Reply classification models~\cite{jayaraman2022combining}. 
In addition to demonstrating that language models leak training data, other efforts focus on understanding the causes for such leakage. 
\citet{jagielski2022measuring} explore the causes of memorization such as training data ordering, \ie, samples can have different privacy risks independent of their content. 
\citet{tirumala2022memorization} study the effect of memorization across variables such as dataset size, learning rate, and model size. 

Related work focus mainly on understanding the leakage in the absence of mitigations. In contrast, we are first to evaluate the interplay of defenses such as PII scrubbing and differential privacy in an end-to-end training pipeline. 
Existing work on training data extraction focuses on \emph{any} type of memorization~\cite{carlini2021extracting} in public pre-trained LMs or models trained from scratch~\cite{kandpal2022deduplicating}, whereas we focus on leakage of PII on fine-tuned LMs given the context where it appears (prefix and suffix), no context, or a list of PII candidates.  

\textbf{Mitigations.}
Several works have proposed solutions to mitigate leakage of private information mainly based on differential privacy (DP) guarantees in the training pipeline. 
\citet{yu2021differentially} and \citet{li2021large} propose an efficient method for differential-privately fine-tuning LMs on private data. 
\citet{shi2021selective} propose selective DP---where DP is only applied to samples containing sensitive information to limit utility degradation.
\citet{stock2022defending} study canary extraction attacks against models fine-tuned from GPT-2 using DP-SGD. 
Closer to our work, \citet{zhao2022provably} propose combining de-duplication, redaction, and DP-SGD to mitigate PII leakage.
It would be most interesting to study how this proposal fares with respect to the risks and metrics we present.

\section{Conclusion}
Our work explores privacy/utility trade-offs of using defenses such as PII scrubbing and Differentially Private training when fine-tuning language models.
We focus on measuring PII leakage from the training data with respect to three different adversary goals: PII extraction, reconstruction, and inference, and provide game-based definitions and leakage metrics for them.
Our findings show that differential privacy is useful in mitigating PII leakage by a large extent but cannot completely eliminate it on its own. 
Traditional data curation approaches such as PII scrubbing are a crucial part of the training pipeline and are still necessary to achieve an appropriate level of protection.
We advocate for the design of less aggressive PII scrubbing techniques that take into account the protection afforded by DP and achieve a better privacy/utility trade-off.

\section*{Acknowledgements}
We are grateful to Victor Rühle, Saurabh Naik, Boris Köpf and the
anonymous reviewers for their suggestions and
feedback that significantly improved this paper.

\bibliographystyle{IEEEtranSN}
\bibliography{bibliography}


\appendices


\section{Datasets}
\label{sec:appendix_datasets}

We survey three datasets that we split into \emph{training}, \emph{validation} and \emph{testing} sets. 
The training and validation sets are equally large and are used to train the target and shadow models respectively. 
We evaluate the perplexity of all models on the testing set, which neither the target nor the shadow models have seen during training. 

\textbf{ECHR.} 
ECHR contains cases from the European Court of Human Rights, which consists of $118\,161$ records. 
Each record consists of $88.12$ tokens on average. 
A case contains a numbered list of \emph{facts}, which are descriptions of the case, such as persons involved or an order of events. 
We split the dataset so that each record contains a single fact. 

Flair NER tags $16\,133$ unique PII of the entity class `person' and we identify that $23.75\%$ of the records contain at least one PII sequence. 
PII is duplicated according to a power law distribution with a mean duplication rate of 4.66. 
A majority of PII ($\geq 90\%$) sequences are duplicated less than 3 times.
For ECHR, many PII from the `person' class are full names (containing a first and a second name), but there are also name abbreviations (\eg, `J.D.' instead of `John Doe').

\textbf{Enron.} 
Enron contains about $600\,000$ real e-mails from 158 employees of the Enron Corporation.
The e-mails were made public by the Federal Energy Regulatory Commission after an investigation.
We observe that e-mails use informal language and may contain typographical errors. 
E-mails do not contain a header (\eg, a `from', `to' or `title' field) unless an e-mail has been forwarded, in which case the forwarded e-mails and their headers are contained in the e-mail's body. 
The e-mails are typically structured with a greeting, followed by the e-mail's content, followed by a footer containing the name and personal information about the sender, such as their e-mail or work address, phone number, and homepage.
We randomly sample a subset of $123\,317$ records from the Enron dataset, where each record consists of $346.10$ tokens on average. 
Flair tags $105\,880$ unique PII sequences from the `person' entity class, and a majority of records contain at least one PII sequence (81.45\%). 
PII sequences have a duplication rate of $11.68$ and contain an average of $3.00$ tokens. 
From qualitative analysis, we observe that PII in the Enron dataset consists of either just the first name or a first and a second name. 

\textbf{Yelp-Health.} 
The Yelp-Health dataset consists of comments made on the \texttt{yelp.com} website about facilities from the `Health \& Medical' category.
Facilities from this category include Psychoanalysts, Dental Hygienists, and Cardiologists among many others.
Compared to ECHR and Enron, Yelp-Health reviews contain relatively short and few PII from the `person' entity class.

We randomly sample a subset of $78\,794$ reviews with an average length of 143.92 tokens per record. 
Flair tags $17\,035$ pieces of PII with an average duplication rate of $5.53$ and $2.17$ tokens per piece. 
In total, $54.55\%$ of records contain at least one PII sequence. 
From a qualitative analysis, we find that pieces of PII from the `person' entity class typically only contain the first name unless they refer to the name of the doctor, who is usually addressed by their last name.

Even though reviews on Yelp-Health were knowingly posted to a public forum\footnote{\url{https://www.yelp.com/developers/documentation/v3/all_category_list}}, these posts may contain sensitive information.
For example, we find detailed descriptions of a relative's disease with timestamps and locations which could be used to re-identify that individual. 
We believe that studying leakage using datasets containing unprocessed data from users who may be unaware of the consequences resulting from re-identification is of particular interest to the community. 

\subsection{PII Definition}
\label{sec:appendix_pii}

This subsection describes what we mean by ``direct" and ``quasi-identifying" PII. 
Our definition is equivalent to \citet{pilan2022text}, who study leakage of PII in large text datasets. 

\begin{itemize}
    \item \emph{Direct identifiers}: A direct identifier is a type of PII that can be used to uniquely identify an individual within a dataset.
    Examples of direct identifiers include an individual's full name, social security number, address of residence, cellphone number or email address. 
    As a result of their sensitive nature, direct identifiers are often subject to legal and regulatory frameworks aimed at safeguarding the privacy of individuals~\cite{gdpr}.
    \item \emph{Quasi-identifiers}: A of PII is a quasi-identifier when it can indirectly identify an individual through its combination with other quasi-identifying information.
    Examples of quasi-identifiers include an individual's age, gender, ethnicity, religion, and occupation. 
    Because of their potential to indirectly identify individuals, both direct and quasi-identifiers are considered personal information and are subject to the same legal and regulatory frameworks in order to protect the privacy of individuals~\cite{gdpr}.
\end{itemize}

\subsection{PII Entity Classes}

We group PII by the following \emph{entity classes}\footnote{\url{https://huggingface.co/flair/ner-english-ontonotes-large}}.

\begin{enumerate}
    \item cardinal: A cardinal value (\eg, ``12").
    \item ordinal: An ordinal value (\eg, ``11th").
    \item date: A date (\eg, ``May 23rd 2015")
    \item event: An event name (\eg, ``Blackhat"). 
    \item fac: A building name (\eg, ``Newark Airport").
    \item gpe: Geo-political entity (\eg, ``UAE").
    \item language: A language name (\eg, ``Catalan")
    \item law: Law names (\eg, ``Bill C-99"). 
    \item money: Currency (\eg, ``1.25m USD")
    \item norp: Affiliation (\eg, ``Alaskan")
    \item person: Names (\eg, ``John Henry Doe")
    \item loc: Location (\eg, ``11th District")
    \item org: Organization (\eg, ``Microsoft")
    \item percent: (\eg, ``(+0.78\%)")
    \item product: (\eg, ``GX-532")
    \item quantity: (\eg ``2000 lbs") 
    \item time: (\eg, ``5:30 pm EDT")
    \item work of art: (\eg, ``Star Wars")
    \item phone number: (\eg, ``(+1)123-456-7890")
    \item email address: (\eg, ``john.doe@anon.com")
    \item url: (\eg, ``www.johndoe.com")
\end{enumerate}

\subsubsection{Replacing PII} 
\label{sec:appendix_replacing_pii}

The attack described in \Cref{alg:estimated_extractability} involves substituting existing PII with PII of the same class.
For example, in the sentence ``Teo Peric is an engineer.” we could replace ``Teo Peric” with a different PII of the class 'person’, resulting in ``Ana Jaksic is an engineer.”. 

\subsection{PII Distribution}
\Cref{tab:dataset_statistic} show the PII counts in each of the three surveyed datasets.

\begin{table*}[htb]
    \centering
    \caption{A summary of the evaluated datasets and statistics for PII from the entity class `person'.
    This table summarizes the part of the dataset that was used to train the model, after splitting the entire dataset into equally large \emph{training} and \emph{validation} sets, and a small \emph{testing} set.}
    \begin{tabular}{@{}lcccccc@{}}
\toprule
            & \multicolumn{1}{l}{Records} & \multicolumn{1}{l}{Tokens / Record} & \multicolumn{1}{l}{Unique PII} & \multicolumn{1}{l}{Records w. PII} & \multicolumn{1}{l}{Duplicates / PII} & \multicolumn{1}{l}{Tokens / PII} \\ \midrule
ECHR        & 118\,161                    & 88.12                            & 16\,133                        & 23.75\%                            & 4.66                                & 4.00                          \\\rowcolor{gray!13}
Enron       & 138\,919                    & 346.10                           & 105\,880                       & 81.45\%                            & 11.68                               & 3.00                          \\
Yelp-Health & 78\,794                     & 143.92                           & 17\,035                        & 54.55\%                            & 5.35                                & 2.17                          \\ \bottomrule
\end{tabular}
    \label{tab:dataset_statistic}
\end{table*}

\subsection{PII Leakage Metrics}
\label{sec:appendix_metrics}
We define formally the metrics introduces in \Cref{sec:metrics}.
\begin{enumerate}
\item \textbf{PII Extractability:} 
We assess PII extractability using precision and recall. 
Precision (also called positive predictive value) quantifies the ratio of true positive predictions to all positive predictions made by a classifier, while recall is the true positive rate, \ie the ratio of true positive predictions to all positive instances in the dataset. 
High recall implies that a significant proportion of PII can be extracted from the LM's generated output, and high precision suggests that a generated PII is highly probable to appear in the training dataset. 
\Cref{alg:estimated_extractability} represents the PII extraction game, and \Cref{formula:succ-extraction} defines the adversary's success as its recall. 
Given the number of unique PII sequences $|\mathcal{C}|$ in $D$, the adversary produces a set of PII sequences $\tilde{\mathcal{C}}$ with a maximum size of $|\mathcal{C}|$.
$\textsc{Precision}$ denotes the expected fraction of correctly identified PII sequences in $\tilde{\mathcal{C}}$ relative to the total number of PII sequences produced by the adversary ($|\tilde{\mathcal{C}}|$).
Recall and precision are formally calculated as follows:
\begin{align}
    \textsc{Recall}    = \EX\left[ \frac{|\mathcal{C} \cap \tilde{\mathcal{C}}|}{|\mathcal{C}|} \right] \quad
    \textsc{Precision} = \EX\left[ \frac{|\mathcal{C} \cap \tilde{\mathcal{C}}|}{|\tilde{\mathcal{C}}|} \right]
\end{align}

\item \textbf{PII Reconstruction and Inference:} 
We measure the top-1 accuracy for PII reconstruction and inference as formalized in \Cref{alg:reconstruction_game}.
Given a randomly sampled sequence $S$ from the training dataset that contains at least one piece of PII $C$, we compute the accuracy of an attacker to produce a guess $\Tilde{C} = C$ for a randomly selected \rectangled{\texttt{[MASK]}} within the sequence.
\begin{align}
    \textrm{Succ}_\textsc{Recon} = \Pr\left[\Tilde{C} = C\right]
\end{align}
The success of the attack depends mildly on the presence of other PII in the sampled sequence $S$. This is because other PII sequences would be scrubbed and the adversary would have less contextual information to make an inference.
\Cref{fig:advanced_reconstruction} illustrates this point for a sequence containing multiple pieces of PII.
\end{enumerate}

\balance

\section{Filling Residual Masks}
\label{sec:fill_masks}

\begin{algorithm}[tpb]
\caption{Fill residual masks}
\label{alg:fill_masks}
\begin{algorithmic}[1]
\Procedure{\textsc{Fill-Masks}}{$S$}
    \State $i \gets 0$
    \While{$\rectangled{\texttt{[MASK]}} \in S$}
        \State $S_i, S \gets \Call{Split}{S, \rectangled{\texttt{[MASK]}}}$
        \Comment{\small At first \rectangled{\texttt{[MASK]}}}
        \State $i \gets i + 1$
    \EndWhile
    \State $S_i \gets S$
    \State $S' \gets S_0$
    \For{$j \gets 1 \textbf{ to } i$}
        \State $w_j \gets \Call{MLM}{S', S_j}$
        \State $S' \gets S' w_j S_j$
    \EndFor
    \Return $S'$ 
\EndProcedure
\end{algorithmic}
\end{algorithm}

\Cref{alg:fill_masks} describes our procedure for filling residual masks in a masked query. 
We use a publicly available masked language model (MLM) to fill in one masked token at a time.
Given a sentence with multiple masked PII, for example ``\rectangled{\texttt{[MASK]}} plays soccer in \rectangled{\texttt{[MASK]}}, England." the goal of \textsc{Fill-Masks} is to fill in all masked tokens in a way that preserves the meaning of the sentence. 

Publicly available MLMs do not jointly fill multiple tokens. (See for instance, \url{https://huggingface.co/docs/transformers/main_classes/pipelines\#transformers.FillMaskPipeline}.)
Hence we resort to filling in masked tokens from left to right using the top token suggested by the MLM. 
In the example above, we would query the MLM twice.
The first query is ``\rectangled{\texttt{[MASK]}} plays soccer in" and the second is ``John plays soccer in \rectangled{\texttt{[MASK]}}, England.", assuming the MLM fills the first mask with ``John". 

Given a sentence $S = S_0 \rectangled{\texttt{[MASK]}} S_1 \ldots \rectangled{\texttt{[MASK]}} S_i$, \cref{alg:fill_masks} first identifies the subsequences $S_0,\ldots,S_i$ (lines 2--6) and then queries the MLM to fill in each masked token separating them from left to right (lines 7--10).
The function $\textsc{MLM}(S_0,S_1)$ queries the MLM on $S_0 \rectangled{\texttt{[MASK]}} S_1$ to predict the most likely token in the position of \rectangled{\texttt{[MASK]}}.


\end{document}